\documentclass[journal]{IEEEtran}
\usepackage[dvips]{graphicx}
\usepackage{lineno,hyperref}
\usepackage{graphicx}
\usepackage{amsmath}
\usepackage{array}
\usepackage{subfigure}
\usepackage{url}
\usepackage{algorithm}
\usepackage{algorithmic}
\usepackage{amssymb}
\usepackage{mathrsfs}
\usepackage{dsfont}  
\usepackage{pifont}
\usepackage{chngpage}  
\usepackage{caption}
\ifCLASSINFOpdf
\else
\fi
\begin{document}
%
\title{Motion-Appearance Interactive Encoding for Object Segmentation in Unconstrained Videos}
%
%
%

\author{Chunchao Guo,
        Jianhuang Lai,
        and Xiaohua Xie\\
        Sun Yat-sen University, China
\thanks{Xiaohua Xie is the corresponding author.}
\thanks{Chunchao Guo, Jianhuang Lai, and Xiaohua Xie are all with the School of Data and Computer Science, Sun Yat-sen University, and with the Guangdong Key Laboratory of Information Security Technology, Guangzhou, 510006, P. R. China. E-mail: chunchaoguo@gmail.com, stsljh@mail.sysu.edu.cn, xiexiaoh6@mail.sysu.edu.cn.}}
\maketitle

\begin{abstract}
We present a novel method of integrating motion and appearance cues for foreground object segmentation in unconstrained videos. Unlike conventional methods encoding motion and appearance patterns individually, our method puts particular emphasis on their mutual assistance. Specifically, we propose using an interactively constrained encoding (ICE) scheme to incorporate motion and appearance patterns into a graph that leads to a spatiotemporal energy optimization. The reason of utilizing ICE is that both motion and appearance cues for the same target share underlying correlative structure, thus can be exploited in a deeply collaborative manner. We perform ICE not only in the initialization but also in the refinement stage of a two-layer framework for object segmentation. This scheme allows our method to consistently capture structural patterns about object perceptions throughout the whole framework. Our method can be operated on superpixels instead of raw pixels to reduce the number of graph nodes by two orders of magnitude. Moreover, we propose to partially explore the multi-object localization problem with inter-occlusion by weighted bipartite graph matching. Comprehensive experiments on three benchmark datasets (i.e., SegTrack, MOViCS, and GaTech) demonstrate the effectiveness of our approach compared with extensive state-of-the-art methods.
\end{abstract}


%
\IEEEpeerreviewmaketitle

\section{Introduction}
%
%
%
%
\IEEEPARstart{T}{he} purpose of video object segmentation is to acquire foreground moving objects in videos.
Foreground object segmentation is greatly significant and has been leveraged for use in various vision tasks, including object appearance modeling \cite{wang2007consensus}, object tracking \cite{colombari2007segmentation}, video matting \cite{hu2013automatic}, activity recognition \cite{brand2000discovery}, and image retrieval \cite{chung2010adaptive}.
Compared with early methods that only consider the case of static camera settings and address this problem through static background subtraction \cite{stauffer1999adaptive, barnich2011vibe}, separating targets in an arbitrary background is inherently more difficult due to camera jittering, motion blur, and the fast and large displacement of targets. Recent years have witnessed much progress \cite{brox2010object,tsai2010motion,zhang2013video,papazoglou2013fast} of handling unconstrained videos, but it remains an open issue that has not yet been adequately explored.


Motion features (e.g., optical flow) and appearance features (e.g., color segmentations) are both important cues for addressing the object segmentation problem with an unconstrained background. However, optical flow generates inaccurate boundaries and it often diffuses under rapid motion, while appearance is severely hindered by cluttered or low-contrast backgrounds.
Therefore, a natural idea is to integrate motion and appearance cues for object segmentation. In many related studies \cite{zhang2013video,papazoglou2013fast,giordano2015superpixel}, the feature-level or decision-level fusion with regard to these two cues has been considered, yet motion and appearance patterns are separately extracted and not integrated in a deeply collaborative way. In other world, traditional manner neglects the intrinsic correlation between motion and appearance patterns. Actually, motion and appearance features for the same target to a certain extent are homologous, and share underlying correlative patterns about object perceptions, including semantic structure, shape, and movement. Therefore, for well detecting a moving target, it is better to exploit the motion and appearance features synergistically rather than to utilize them separately. Along this line, we develop an interactively constrained encoding (ICE) approach for integrating motion and appearance cues and incorporate it into a coarse-to-fine framework.
The procedure of feature encoding is interactive between multi-type cues; that is, ICE imposes motion constraints during appearance feature encoding, and vice versa.
Unlike many existing methods in which multiple-feature fusion only serves the initialization stage, our method performs ICE throughout all of the stages of the coarse-to-fine framework.
Especially, ICE allows our method to capture the semantic structure of object perception while refining moving regions.

\begin{figure*}[!htb]
\centering
\hspace{-1in}
\includegraphics[width=0.7\textwidth]{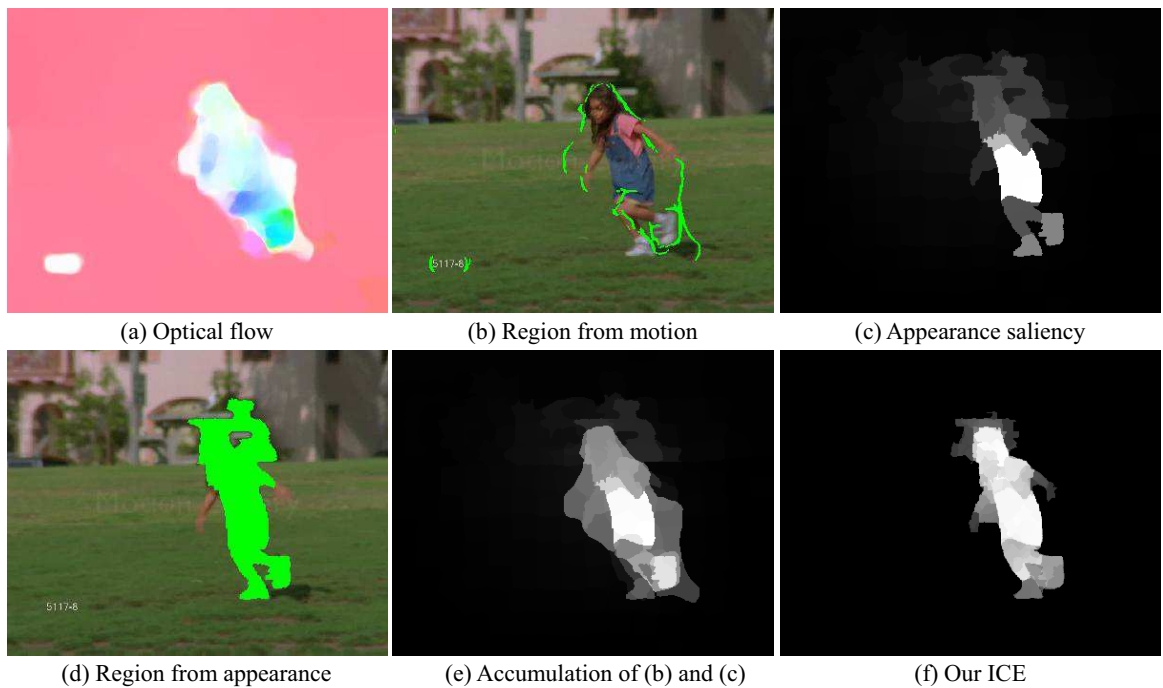}
\caption{Illustration of separate encoding and Interactively Constrained Encoding (ICE) of motion and appearance patterns for a moving target. Best view in color.}
\label{figMotivation}
\end{figure*}

Figures \ref{figMotivation}(f) and \ref{figJointEncoding}(d) illustrate living examples of ICE and Figure \ref{figFlowchart} demonstrates our framework.
In Figure \ref{figMotivation}, (b) and (c) are the segmentation cues extracted by \cite{papazoglou2013fast} and \cite{yang2013saliency}, respectively.
Figure \ref{figMotivation}(e) is the combination of (b) and (c), but the motion and appearance cues are encoded separately.
By considering the interactively constrained encoding on two cues, our approach obviously gets more uniform values inside the target than (c) and (e), and preserves much more accurate boundaries than (b) and (e). Overall, ICE exhibits more potential in high-level object perception.


Besides putting forward ICE, we employ another two strategies in addressing the video object segmentation problem. First, the superpixel representation is used to reduce the computation complexity. Because our method performs ICE throughout the whole framework, the superpixel-level graph still works well and maintains the ability of perceiving targets in the video even with a cluttered background. Second, to handle the multi-object initialization with inter-occlusion, we propose to activate maximum bipartite graph matching between adjacent frames at the proposal level, which re-assigns coarse IDs to different occluded objects. It should be noted that the inter-occlusion is often ignored by previous studies.

%

The remainder of this paper is structured as follows. Related works are reviewed in Section 2. Section 3 presents our approach. Section 4 demonstrates the results on three benchmark datasets, and our conclusions are drawn in Section 5.

\section{Related Works}
This paper aims to detect and segment moving foreground objects in videos, a goal shared by previous works on background subtraction and video object segmentation.
Related works are briefly introduced in this section.

\subsection{\textbf{Background Subtraction}}
When detecting moving regions, a way off the shelf is to model pixel-wise backgrounds and then subtract it to find pixels whose differences exceed a threshold. Both parametric \cite{stauffer1999adaptive,wren1997pfinder,sheikh2005bayesian} and nonparametric \cite{barnich2011vibe} mechanisms can be adopted for constructing backgrounds.
The single Gaussian model \cite{wren1997pfinder} is a simple method for fitting the distribution of pixel intensity, but it is insufficient in complex scenes.
The elaborate mixture of Gaussian \cite{stauffer1999adaptive} provides a better distribution of background pixels and it is usually eligible for non-cluttered scenes.
Chen et al. \cite{chen2007efficient} proposed a hierarchical block-based approach that combines Mixture of Gaussian and a contrast histogram.
With respect to nonparametric approaches, the background model in \cite{barnich2011vibe} is built upon a set of background samples and is updated by randomly selecting samples. \cite{barnich2011vibe} also considers neighbor pixels during updating, which is different from Gaussian mixture models \cite{stauffer1999adaptive}.
Nevertheless, the common drawback of traditional background modeling approaches is that they usually require static camera settings, and thus cannot cope well with moving backgrounds.

\subsection{\textbf{Video Object Segmentation}}
Video object segmentation has emerged as a feasible solution for tackling arbitrary background. Conventional methods of video object segmentation can be categorized into the supervised mode and the unsupervised mode.
\subsubsection{Supervised Methods}
Supervised approaches usually require manual annotations in several key frames to explicitly indicate moving objects.
Tsai et al. \cite{tsai2010motion} proposed to track human-labeled regions and segment the remaining frames utilizing a multi-label Markov random field model.
Chockalingam et al.\cite{chockalingam2009adaptive} also requires a manual label that indicates the location of a moving target.
Recently, some methods \cite{fu2014object,wang2014video} have focused on co-segmenting objects, given multiple videos where the same targets appear simultaneously.
Those can be categorized as weakly supervised tasks, as the same video object is must be present in multiple videos and the co-occurrence hints at what the object looks like.

\subsubsection{Unsupervised Methods}
Unsupervised segmentation methods mainly build on motion trajectories or optimization in a graph with integrating multiple cues.

\paragraph{Methods based on Motion Trajectory} Trajectories provide a natural way to describe object movement. Those methods usually obtain trajectories by linking dense points or objects, and then the point trajectories are clustered and mapped into pixel labels.
Brox et al. \cite{brox2010object} exploited dense flow points to cluster long-term motion trajectories, and the temporally consistent clusters alleviated the influence of objects that were sometimes static in the sequence. Fragkiadaki et al. \cite{fragkiadaki2012video} detected discontinuities for trajectory embedding to obtain motion boundaries, and thus segment objects from world scenes.
The common underlying assumption that supports these methods is motion homogeneity, where the points within objects share a single similarity transformation. However, deformable objects apparently do not meet this requirement and may lead to poor performance. Moreover, these methods rely heavily on the reliability of optical flow and explore insufficient appearance clues.

\paragraph{Methods based on Coarse-to-Fine Graph Optimization}
Beyond tracking dense points, an alternative that benefits more from both motion and appearance is the coarse-to-fine optimization scheme in a graph. Raw objects are initially localized with the help of optical flow \cite{papazoglou2013fast} or generic object proposals \cite{zhang2013video}, and then multiple clues are incorporated into an optimization procedure to refine the foreground labels.
Lim et al. \cite{lim2012modeling} employed block-wise density propagation to obtain likelihood maps, and then optimized those maps. However, this approach involves a large number of parameters and thus is not feasible for generalization. Wang et al. \cite{wang2015saliency} designed a framework that incorporates robust geodesic measurement to segment video targets.
\cite{cao2016unsupervised} optimizes a weighted graph at the pixel level using the shortest path algorithm.
Giordano et al. \cite{giordano2015superpixel} incorporated properties of a compact geometrical structure and optimized the graph around each moving region based on appearance and perceptual organization.

Although motion and appearance patterns are both used in above-mentioned methods when optimizing a graph, the common drawback of these methods is that motion and appearance are treated as isolated components at a low level without considering their homologous property.
Furthermore, existing methods often ignore the inter-occlusion problem in multiple object location. Our method tends to address all these shortcomings.


\section{The Proposed Approach}

\subsection{\textbf{Overview and Notations}}
\begin{figure*}[!htbp]
\centering
\captionsetup{justification=centering}
\hspace{-1.2in}
\includegraphics[width=0.8\textwidth]{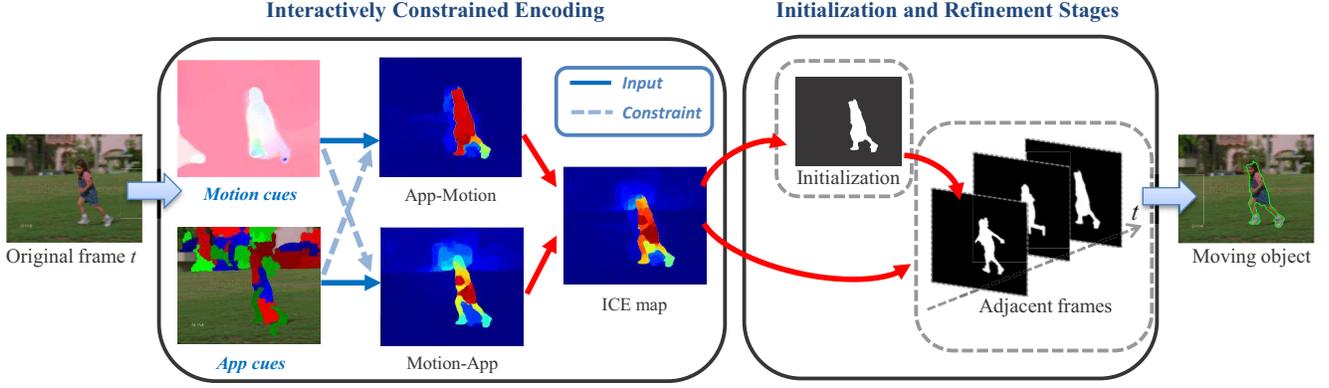}
\caption{Flowchart of the framework.}
\label{figFlowchart}
\end{figure*}

Our framework comprises two stages: the label initialization stage and the refinement stage. The proposed interactively constrained encoding (ICE) will be used throughout both stages. Figure \ref{figFlowchart} briefly illustrates our approach.

For clarity, the notations used in our paper are provided in Table \ref{table:notaions}.

\begin{table}[htbp]
\centering
\begin{tabular}{l|p{0.25\textwidth}}  
\hline
Symbol                                            &Definition    \\ \hline  
$F^{i} \in \mathcal{X}^{W\times H}$               & The $i$th frame with width $W$ and height $H$ \\
$\mathcal{F}= \left \{ F^{i} \right \}_{i=1}^{K}$   & A video sequence with $K$ frames  \\
$\left ( j,k \right )$                            & A pixel index  \\
$O^{i}$                                          & The two-channel optical flow image \\
$V^{i}$                                        &The intensity magnitude of $O^{i}$ \\
$E^{i}$,                                         &The gradient magnitude of $O^{i}$ \\
$C^{i}$                     & The three-channel color optical flow image \\
$Y_{RGB}^{i}$     & The appearance saliency map of $F^{i}$ \\
$Y_{C}^{i}$     & The appearance saliency map of $C^{i}$ \\
$\mathcal{P}_{RGB}^{i}= \left \{ P_{RGB}^{i,r} \right \}_{r=1}^{N_{1}}$   & A set of object proposals in $F^{i}$ \\
$\mathcal{P}_{C}^{i}= \left \{ P_{C}^{i,r} \right \}_{r=1}^{N_{2}}$    & A set of object proposals in $C^{i}$ \\
$P^{i,r}$      & The $r$th proposal in $F^{i}$ or $C^{i}$ \\
$D^{i,r}$      & The binary mask of $P^{i,r}$  \\
$B^{i,r}$      & The binary boundary/edge map of $P^{i,r}$  \\
$G$            & The gradient/boundary strength for a proposal $P^{i,r}$  \\
$I$            & The intensity strength for a proposal $P^{i,r}$  \\
$G^{i}$            & The accumulated gradient/boundary strength of $\mathcal{P}^{i}$ in $F^{i}$ or $C^{i}$ \\
$I^{i}$            & The accumulated intensity strength of $\mathcal{P}^{i}$ in $F^{i}$ or $C^{i}$  \\
$T^{i}$            & The trimap for $C^{i}$ \\
$\textbf{M}^{i}$   & The ICE map for $F^{i}$  \\
$H_{p}$               & Multiple concatenated histograms for a superpixel/node $p$  \\
$\mathcal{D}$       & The Bhattacharyya distance  \\ \hline
\end{tabular}
\vspace{0.1in}
\caption{The notations in this paper.} \label{table:notaions}
\end{table}

\subsection{\textbf{Interactively Constrained Encoding (ICE)}}

\subsubsection{Overview of ICE}
To exploit the homologous properties of multimodal cues and capture semantic structural information on object perception, appearance restrictions are leveraged to induce the extraction of motion patterns, and vice versa.

The motion cues used in our approach mainly include optical flow \cite{brox2011large} and its further derivative features, such as a color image and a gradient map of optical flow. In terms of appearance, object proposals \cite{krahenbuhl2014geodesic} are the primary cues in our approach. Appearance saliency \cite{yang2013saliency}, trimap and color descriptors are also exploited as supplementary cues.

The gradient magnitude map $E^{i}$ in the optical flow field ${O}^{i}$ is calculated by:
\begin{equation}\label{flowgrad}
    E^{i} = \left \|  \nabla O^{i} \right \|_{2}.
\end{equation}

For $P^{i,r}$, its gradient or boundary strength $G$ in the optical flow field is formulated as
\begin{equation}\label{fgrad}
    G\left ( P^{i,r} \right )= \frac{1}{Z^{i,r}_{G}}\sum_{j=1}^{H}\sum_{k=1}^{W} E^{i}_{j,k} \cdot \mathds{1} \left ( B^{i,r}_{j,k}=1 \right ),
\end{equation}
where $\left ( j,k \right )$ is a pixel index, $\mathds{1}$ is the indicator function, and $Z^{i,r}_{G}=\sum_{j=1}^{H}\sum_{k=1}^{W} B_{j,k}^{i,r}$ is a normalization factor. Given $P^{i,r}$, Eq.\eqref{fgrad} evaluates the strength of a proposal boundary that overlaps with the boundary of a moving object.

In the same way, intensity strength $I$ in the optical flow field for a proposal $P^{i,r}$ is
\begin{equation}\label{flowinten}
    I\left ( P^{i,r} \right )= \frac{1}{Z^{i,r}_{I}}\sum_{j=1}^{H}\sum_{k=1}^{W} V^{i}_{j,k} \cdot \mathds{1} \left ( D^{i,r}_{j,k}=1 \right ),
\end{equation}
where $Z^{i,r}_{I}=\sum_{j=1}^{H}\sum_{k=1}^{W} D_{j,k}^{i,r}$ is also a normalization factor. Given $P^{i,r}$, Eq.\eqref{flowinten} assesses its intensity strength weighted by the intensity of the optical field. This metric also indicates the likelihood of belonging to moving targets for $P^{i,r}$.

\subsubsection{Appearance-Constrained Motion}
We adapt appearance restrictions to the motion field; that is, we aim to encode patterns in $O^{i}$, by following the manner of extracting appearance features.
Given $C^{i}$, we calculate its appearance saliency map \cite{yang2013saliency} $Y_{C}^{i}$, its color name descriptors \cite{van2009learning} and its trimap $T^{i}$. The construction of $T^{i}$ is individually introduced later. Color name descriptors categorize each pixel into the eleven semantic color names and thus produce a histogram with eleven dimensions.

Then, we rank and accumulate $\mathcal{P}_{C}^{i}= \left \{ P_{C}^{i,r} \right \}_{r=1}^{N_{2}}$ in ${C}^{i}$ based on Eqs.\eqref{fgrad} and \eqref{flowinten}.
The accumulated boundary or gradient strength of $\mathcal{P}_{C}^{i}$ is
\begin{equation}\label{cflowgrad}
  G_{C}^{i}=\sum_{r=1}^{N_{2}}B_{C}^{i,r} \cdot G\left ( P_{C}^{i,r} \right )
\end{equation}

Given $B_{C}^{i,r}$, as a mask for $P_{C}^{i,r}$, Eq.\eqref{cflowgrad} assigns a uniform value $G\left ( P_{C}^{i,r} \right )$ for each pixel belonging to $P_{C}^{i,r}$. Considering that those proposals are generated based on similarity in $C^{i}$ and are thus compact in motion, the settings in Eq.\eqref{cflowgrad} can preserve spatial layouts of targets when many interior boundaries occur inside them.

In the same way, the accumulated intensity strength of $\mathcal{P}_{C}^{i}$ is
\begin{equation}\label{cflowinten}
  I_{C}^{i}=\sum_{r=1}^{N_{2}}B_{C}^{i,r} \cdot I\left ( P_{C}^{i,r} \right ),
\end{equation}

Note that, while appearance constraints were imposed during the above feature encoding, the whole procedures are conducted in optical flow field $O^{i}$ or $C^{i}$. Thus, $Y_{C}^{i}$, $G_{C}^{i}$, $I_{C}^{i}$, $T^{i}$ and the color descriptors of $C^{i}$ are categorized into the appearance-constrained motion.

\paragraph{Trimap} 
A trimap $T^{i}$ denotes the division of definite foreground, definite background, and ambiguous regions in $F^{i}$.
During model optimization, $T^{i}$ reveals the correlations of regions both in a local and global view, and thus can narrow down our focus to only ambiguous regions.

Unlike \cite{kim2014salient}, where the goal is to find salient objects in appearance, we aim for localizing moving areas. Hence, our moving trimap $T^{i}$ is built upon $Y_{C}^{i}$.
$Y_{C}^{i}$ is first subdivided into equal-sized blocks under three different spatial scales, with the number of blocks in each scale being $2 \times 2$, $3 \times 3$, and $4 \times 4$, respectively.
Similar to the settings in \cite{kim2014salient}, the Otsu's algorithm \cite{otsu1975threshold} with seven-level threshold is individually applied to each block in each scale.
Then, all three maps are summed to global map ${Y^{i}}'$. Hence, ${Y^{i}}'$ is a map with 21 levels. $T^{i}$ is then constructed by globally thresholding ${Y^{i}}'$ using Eq.\eqref{trimap}, through a conservative scheme to ensure the purity of the definite foreground:
\begin{equation}\label{trimap}
   T^{i}\left ( j,k \right )=\left\{\begin{matrix}
    1, & if  & {Y^{i}}' \left ( j,k \right )\geq \theta_{1},\\
    0,& if & {Y^{i}}' \left ( j,k \right )\leq \theta_{2}, \\
    0.5,&otherwise
\end{matrix}\right.
\end{equation}
where $\left ( j,k \right )$ is a pixel index in frame $F^{i}$. In this work, $\theta_{1}$ and $\theta_{2}$ are set as 18 and 6, respectively.

\begin{figure*}[tb]
\centering
\hspace{-1.2in}
\includegraphics[width=0.8\textwidth]{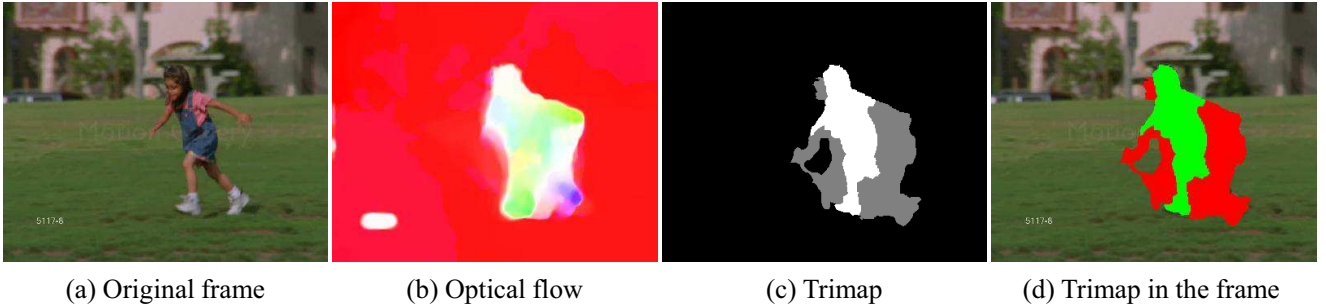}
\caption{Illustration of a trimap. In (d), the green middle region represents definite foreground, the red color marks ambiguous regions, and the remaining region is the background.}
\label{figTrimap}
\end{figure*}

An example of trimap in Figure \ref{figTrimap} indicates that it benefits for localizing objects and narrowing down our attention region.

Thus, an appearance-constrained motion map $\textbf{M}^{i}_{C}$ is expressed as
\begin{equation}\label{appconstrainmotion}
\textbf{M}^{i}_{C}=G_{C}^{i} + I_{C}^{i} + \alpha \cdot Y_{C}^{i} + \beta \cdot T^{i},
\end{equation}
where $\alpha$ and $\beta$ are set as 0.9 and 0.5 in our experiments, respectively.
Each term in Eq.\eqref{appconstrainmotion} is restricted by the object-level appearance relations in $C^{i}$.

\subsubsection{Motion-Constrained Appearance}
In contrast to $\textbf{M}^{i}_{C}$, which has the aid of appearance constraints, motion restrictions can also be leveraged to encode appearance patterns. Object proposals in color space uncover the potential of maintaining the semantic structure even in cases of optical flow failure. We first rank and accumulate $\mathcal{P}_{RGB}^{i}$ based on Eqs.\eqref{fgrad} and \eqref{flowinten}.


The accumulated gradient strength of $\mathcal{P}_{RGB}^{i}$ is
\begin{equation}\label{rgbgrad}
  G_{RGB}^{i}=\sum_{r=1}^{N_{1}}B_{RGB}^{i,r} \cdot G\left ( P_{RGB}^{i,r} \right ).
\end{equation}
Given $B_{RGB}^{i,r}$, Eq.\eqref{rgbgrad} assigns a uniform value $G\left ( P_{RGB}^{i,r} \right )$ for each pixel belonging to $P_{RGB}^{i,r}$.
Here optical flow boundaries $E^{i}$ must be pre-processed before using Eq.\eqref{rgbgrad} is required to avoid many zero values. Given that the true image boundaries in $F^{i}$ are slightly misaligned with $E^{i}$ in $O^{i}$, we first produce an expansion map ${E^{i}}'$ by dilating $E^{i}$ to make it overlap with the boundaries of $F^{i}$ and $\mathcal{P}_{RGB}^{i}$.

In the same way, the accumulated intensity strength of $\mathcal{P}_{RGB}^{i}$ is
\begin{equation}\label{rgbinten}
  I_{RGB}^{i}=\sum_{r=1}^{N_{1}}B_{RGB}^{i,r} \cdot I\left ( P_{RGB}^{i,r} \right ).
\end{equation}

Note that, while optical flow cues aid feature encoding here, the whole procedures are still operated on $\mathcal{P}_{RGB}^{i}$. $\mathcal{P}_{RGB}^{i}$ belongs to appearance patterns, as they are generated from RGB frames and the generation is uninfluenced by optical flow. Thus, the feature maps in this part are categorized into the motion-constrained appearance.

The motion-constrained appearance map $\textbf{M}^{i}_{RGB}$ is then formulated as
\begin{equation}\label{motionconstrainapp}
\textbf{M}^{i}_{RGB}=G_{RGB}^{i} + I_{RGB}^{i}+\alpha \cdot Y_{RGB}^{i}.
\end{equation}


\subsubsection{The Eventual ICE Map}
Having achieved the motion-constrained appearance and the appearance-constrained motion, we assign equal weights to the two terms.
An ICE map is then calculated by
\begin{equation}\label{ICE}
  \textbf{M}^{i}= \textbf{M}^{i}_{C}+\textbf{M}^{i}_{RGB}.
\end{equation}
We then normalize $\textbf{M}^{i}$ to $\left [0,1 \right ]$. Eventually, $\textbf{M}^{i}$ indicates the probabilities of being a moving foreground for the pixels in $F^{i}$.

\begin{figure}[tb]
\centering
\includegraphics[width=0.5\textwidth]{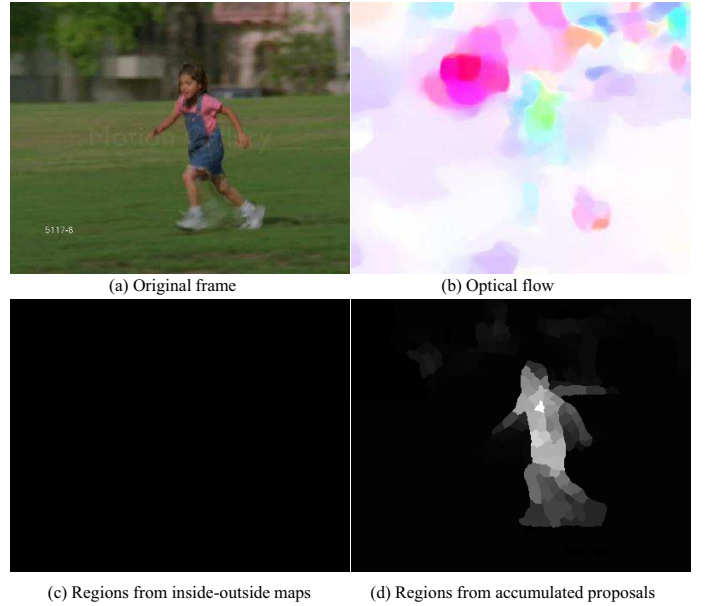}
\caption{Illustration of moving object perception by different method. Optical flow in (b) is inaccurate due to rapid motion.
The initial moving region in (c) is calculated using inside-outside maps according to \cite{papazoglou2013fast}, which only relies on optical flow and fails in this case. (d) is our accumulated object proposals induced by optical flow. Motion(optical flow) with appearance (RGB proposals) release potentials of object perception due to the homologous property.}
\label{figJointEncoding}
\end{figure}

Our ICE scheme considers mutual restrictions from multimodal cues for the same target, and it tends to improve the robustness and accuracy of perceiving moving targets.
Figure \ref{figJointEncoding} provides a living example, which shows that our approach can work effectively in variant environments, even in which the optical flow method fails.

\subsection{\textbf{Label Initialization}}
Having obtained an ICE map $\textbf{M}^{i}$, an adaptive threshold $t^{i}$ is computed to binarize $\textbf{M}^{i}$.
$t^{i}=0.5 \cdot \left ( \mu \left (  \textbf{M}^{i} \right )+\rho  \left (  \textbf{M}^{i} \right ) \right )$, where $\mu \left ( \cdot \right )$ is the average value of $\textbf{M}^{i}$ and $\rho \left ( \cdot \right )$ denotes the threshold computed on $\textbf{M}^{i}$ using Otsu's algorithm \cite{otsu1975threshold}. The initial foreground or background map $X^{i}$ is obtained by
\begin{equation} \label{binary}
X^{i}\left ( j,k \right )=\left\{\begin{matrix}
1, & if & \textbf{M}^{i}\left ( j,k \right ) \geq t^{i} \\
0,& otherwise  &
\end{matrix}\right .
\end{equation}
where $\left ( j,k \right )$ is a pixel index in $F^{i}$. Eq.\eqref{binary} usually works well in cases of single-object segmentation and multi-object segmentation without occlusions.


Nevertheless, for a sequence $\mathcal{F}= \left \{ F^{i} \right \}_{i=1}^{K}$ containing multiple object movement, there may exist inter-occlusion in certain frames $\left \{ F^{o+1},..,F^{o+r}\right \}$ after initialization.
To partially alleviate this problem, we introduce a conservative procedure to activate the inter-occlusion handling.
A decrease in the number of detected targets suggests that certain targets leave the scene or inter-occlusion happens. Hence, our decision criterion is that the number of initial targets decreases from $n$ to $m$ and then returns to $n$.
Here, $n$ is the number of targets before decreasing, and $m$ is the number of detected occluded blobs, s.t. $m<n$.
This criterion indicates that inter-occlusion occurs from $F^{o+1}$ and the occluded blobs re-split to $n$ isolated targets after $F^{o+r}$.
Thus, the $m$ occluded blobs in $\left \{ F^{o+1},..,F^{o+r}\right \}$ can be split into proposals and re-assigned IDs.

We construct a graph $\mathcal{G}_{1}=\left(\mathcal{V}_{1},\mathcal{E}_{1} \right)$ for every two adjacent frames $F^{i},F^{i+1}$, where $\mathcal{V}_{1}=\mathcal{P}_{RGB}^{i} \cup \mathcal{P}_{RGB}^{i+1}$ and $\mathcal{P}_{RGB}^{i+1}$ is the set of proposals inside the occluded blobs.
Given that $\mathcal{P}_{RGB}^{i}$ and $\mathcal{P}_{RGB}^{i+1}$ are disjoint sets of proposals from adjacent frames $F^{i}$ and $F^{i+1}$, respectively, $\mathcal{G}_{1}$ naturally consists of a bipartite graph. Every edge $e\in \mathcal{E}_{1}$ indicates the cost between two nodes/proposals from $P_{RGB}^{i,j}$ and $P_{RGB}^{i+1,k}$, respectively.
The cost of $e$ is calculated by three terms, including the Bhattacharyya distance between concatenated of simple RGB and LAB histograms, the normalized difference of the sizes of boundingboxes, and the normalized centroid distance between $P_{RGB}^{i,j}$ and $P_{RGB}^{i+1,k}$.
The maximum matching of bipartite graph $\mathcal{G}_{1}$ is solved by the Hungarian algorithm.




\subsection{\textbf{Label Refinement}}
The purpose of label refinement is to reduce the misclassifications generated by local nature during foreground initialization, particularly near the edges of moving targets. An energy minimization formulation is thus introduced to enforce spatial consistency of targets.
Given that superpixels \cite{achanta2012slic} are usually achieved through a conservative strategy to ensure highly local compactness, our framework advocates them as basic units in the spatiotemporal graph $\mathcal{G}$.
This setting, an operation at a middle level, may sacrifice some accuracy compared with pixel-by-pixel optimization, but it is still feasible and faster than pixel-wise labeling because the number of nodes in $\mathcal{G}$ has been reduced by two orders of magnitude.
Here, our ICE is still used in unary potentials to maintain the ability to capture more structural information about object perception.

Given a video sequences $\mathcal{F}= \left \{ F^{i} \right \}_{i=1}^{K}$,
we formulate $\mathcal{F}$ as a spatiotemporal graph $\mathcal{G}=\left ( \mathcal{S}, \mathcal{E} \right )$,
with an initial node label set $\mathcal{L}=\left \{ l_{p} \right \}_{p=1}^{N}$, $l_{p} \in \left \{ 0,1 \right \}$.
Here, $\mathcal{S}$ is the collection of superpixels/nodes and $\mathcal{E}$ is the set of edges.

Then the energy in the spatiotemporal graph $\mathcal{G}$ is defined as
\begin{equation}\label{energy}
\begin{aligned}
  \textbf{E}\left ( \mathcal{L} \right )&=\sum_{p\in \mathcal{G}} \mathcal{U}_{p}\left ( l_{p} \right )+ \lambda_{1} \sum_{p\in \mathcal{G}}\sum_{q\in \mathcal{N}_{s} \left ( p \right )} \mathcal{V}_{pq} \left ( l_{p},l_{q} \right ) \\
  &+ \lambda_{2} \sum_{p\in \mathcal{G}}\sum_{r\in \mathcal{N}_{t} \left ( p \right )} \mathcal{V}_{pr} \left ( l_{p},l_{r} \right )\\
  &s.t. \ \ \ l_{p},l_{q},l_{r} \in \left \{ 0,1 \right \},
\end{aligned}
\end{equation}
where $\mathcal{N}_{s}\left ( p \right )$ and $\mathcal{N}_{t}\left ( p \right )$ represent the spatial neighbors and temporal neighbors of node $p$, respectively.
All of the superpixels in the same frame that spatially connected to node $p$ consist of $N_{s}\left ( p \right )$, whereas all of the superpixels in the forward or backward adjacent frames that overlap with $p$ comprise $N_{t}\left ( p \right )$.

There exist three terms in Eq.\eqref{energy}. Unary potential $\mathcal{U}_{p}$, also called a data term, represents the likelihood of node $p$ belonging to label $0$ or $1$. Pairwise potential $\mathcal{V}$, also called a smooth term, includes spatial pairwise potential $\mathcal{V}_{pq}$ in the same frame and temporal pairwise potential $\mathcal{V}_{pr}$ across adjacent frames.
Below we detail the definitions of $\mathcal{U}$ and $\mathcal{V}$.

\subsubsection{Unary Potential $\mathcal{U}$}
The first term in unary potentials is calculated using the ICE maps $\mathcal{M}=\left \{ \textbf{M}^{i} \right \}_{i=1}^{K}$.

The ICE map for node $p$ is transformed into unary values by
\begin{equation}\label{dataTerm2}
    \mathcal{U}_{1}\left ( l_{p} \right )=\left\{\begin{matrix}
     -log\left ( 1- \textbf{M}^{i} \left ( p \right ) \right )& if & l_{p}=0,  \\
     -log\left ( \textbf{M}^{i} \left ( p \right ) \right ) &  otherwise
    \end{matrix}\right.
\end{equation}
where $\textbf{M}^{i} \left ( p \right )$ is the normalized summation value of node $p$ in $\textbf{M}^{i}$.
The second data term, Eq.\eqref{dataTerm2}, aims to penalize the initial background superpixels that own large likelihood values in ICE, and the initial foreground nodes with small values in ICE.

As complementary, the weighted histograms $\mathcal{H}=\left \{ H_{p} \right \}_{p=1}^{N}$ from each superpixel/node $p$ are also employed.
$H_{p}$ is a feature pool that concatenates three types of histograms:
the Bag of Words (BoW) histograms respectively projected by a dense SIFT dictionary and an RGB value dictionary, color name descriptors as described in the appearance-constrained motion, and the concatenation of simple histograms in four color spaces including RGB, HSV, LAB, and YCbCr.
Given that histograms of node $p$ may be sparse, we employ the $k$ nearest spatial neighbors of $p$ to weight it through a Gaussian kernel and achieve a weighted $H_{p}$.

\begin{equation}\label{dataTerm1}
\mathcal{U}_{2}\left ( l_{p} \right )=\left\{\begin{matrix}
     1- \mathcal{D} \left(H_{p},H\left({Q}_{fg}\right) \right) & if & l_{p}=0,  \\
     1- \mathcal{D} \left(H_{p},H\left({Q}_{bg}\right) \right) & otherwise
\end{matrix}\right.
\end{equation}
where ${Q}_{bg}$ and ${Q}_{fg}$ are sets of initial background and foreground superpixels, respectively. $\mathcal{D}$ represents the Bhattacharyya distance of the feature vectors between node $p$ and ${Q}_{fg}$ or ${Q}_{bg}$. The purpose of formulating the first data term as Eq.\eqref{dataTerm1} is to assign a higher unary value to background superpixels that are close to foreground objects after feature embedding, and vice versa.

Thus, the unary potential for node $p$ is
\begin{equation}\label{dataTerm}
   \mathcal{U}_{p}\left ( l_{p} \right )= \mathcal{U}_{1}\left ( l_{p} \right ) +\mathcal{U}_{2}\left ( l_{p} \right ).
\end{equation}


\subsubsection{Pairwise Potential $\mathcal{V}$}

To calculate the degree of agreement between two spatial or temporal adjacent nodes, we need to define feature representation and metrics for nodes.
Two types of cues are used here to compute the pairwise potential: histograms $H$ and boundary connectivity $\mathcal{C}$.
$H$ demonstrates the appearance and motion similarity, and $\mathcal{C}$ depicts the spatial closeness of adjacent nodes.

According to visual perception, two superpixels/nodes $p$ and $q$ are likely to be intimate and compact if they connect much with each other.
In this case, a large percentage of boundary pixels for $p$ and $q$ are overlapped.
The boundary connectivity value $\mathcal{C}$ is thus defined as
\begin{equation}\label{connect}
    \mathcal{C} \left(p,q \right)=\left\{\begin{matrix}
0, & if  &  l_{p}=l_{q} \\
\frac{Len\left(p \right) \cap Len\left(q \right)}{ min \left( Len\left(p \right), Len\left(q \right) \right) } & if &  l_{p}\neq  l_{q}
\end{matrix}\right.
\end{equation}
where $Len\left(p \right)$ denotes the perimeter of superpixel $p$ and $Len\left(p \right) \cap Len\left(q \right)$ is the overlapped length of their perimeters.

Given two nodes $p$ and $q$, s.t. $q\in \mathcal{N}_{s} \left ( p \right )$, spatial pairwise potential $\mathcal{V}_{pq}$ is written as
\begin{equation}\label{spatialpair}
   \mathcal{V}_{pq}\left ( l_{p},l_{q} \right )=\left\{\begin{matrix}
    0, & if  & l_{p}= l_{q}\\
    1- \mathcal{D} \left(H_{p},H_{q} \right) + \mathcal{C} \left(p,q \right),& if & l_{p} \neq l_{q}
\end{matrix}\right.
\end{equation}

According to node $p$ and $r\in \mathcal{N}_{t} \left ( p \right )$, temporal pairwise potential $\mathcal{V}_{pr}$ can be expressed as
\begin{equation}\label{temppair}
   \mathcal{V}_{pr}\left ( l_{p},l_{r} \right )=\left\{\begin{matrix}
    0, & if  & l_{p}= l_{r}\\
    1- \mathcal{D} \left(H_{p},H_{r} \right) ,& if & l_{p} \neq  l_{r}
    \end{matrix}\right.
\end{equation}
Eqs.\eqref{spatialpair} and \eqref{temppair} aim to penalize adjacent nodes that are assigned with different initial labels.

Based on Eqs.\eqref{energy}, \eqref{dataTerm}, \eqref{spatialpair}, and \eqref{temppair}, the refined foreground labels are achieved by minimizing the objective function:
\begin{equation}\label{miniEnergy}
   \mathcal{L}^{\ast }=\underset{\mathcal{L}}{arg\mathit{min} \textbf{E} \left ( \mathcal{L} \right )}.
\end{equation}
Given that the unary and pairwise potentials in our approach are submodular, this task can be done via the graph cut algorithm.

%
%
%

\section{Experiments}

\subsection{\textbf{Experimental Settings}}

\paragraph{\textbf{Datasets}}
Three benchmark datasets were employed to evaluate our method: SegTrack \cite{tsai2010motion}, MOViCS \cite{chiu2013multi}, and GaTech \cite{grundmann2010efficient}.
\textbf{SegTrack} \cite{tsai2010motion} is a commonly used dataset for video object segmentation. It contains six video sequences named \emph{birdfall, cheetah, girl, monkeydog, parachute,} and \emph{penguin}. A pixel-level ground-truth is provided for the primary foreground object in each video. We follow the same criterion as in \cite{zhang2013video,papazoglou2013fast,lee2011key}, where the penguin video is discarded because only one penguin is annotated among a group of penguins.
\textbf{MOViCS} \cite{chiu2013multi} was initially proposed for video object co-segmentation. It is a weakly supervised pipeline for segmenting objects in multiple relevant videos. The dataset has four video sets: \emph{chickensAll, lionsAll, giraffesAll and tigersAll}. Each video set contains two to four videos. In each video, the authors provide the ground truth of object class labeling for five frames that are equidistantly sampled from the video.
The \textbf{GaTech} video segmentation dataset \cite{grundmann2010efficient} consists of 15 sequences, and the video length ranges from 1 second to 28 seconds.

\paragraph{\textbf{Evaluation Metrics}}
With respect to quantitative analysis, the popular average per-frame pixel error \cite{tsai2010motion} was used as a basic metric. The metric is defined as
\begin{equation}\label{avgPixError}
    error=\frac{XOR\left ( FG,GT \right )}{K},
\end{equation}
where $FG$ is the labeling results for all frames output by the segmentation approaches, $GT$ is the ground-truth labels, and $K$ is the total number of frames for a sequence. Given that the average per-pixel error is an absolute quantitative metric and can vary in a wide range influenced by video resolution, we supplement a normalized metric named average labeling precision
\begin{equation}\label{avgPrecision}
    precision = 1-\frac{XOR\left ( FG,GT \right )}{K\cdot N_{0}},
\end{equation}
where $N_{0}$ is the total number of pixels in each frame.

\paragraph{\textbf{Implementation Details}} All of the experiments were run on a computer with Intel Core i7(3.4GHz) and 8GB RAM. In our model, SLIC superpixels \cite{achanta2012slic} were used with a regularizer 0.1 and a regionsize 20.
The GOP algorithm \cite{krahenbuhl2014geodesic} with default parameters was used to generating proposals due to its fast computation.
The $\lambda_{1}$ and $\lambda_{2}$ in Eq.\eqref{energy} were set to 3 and 2, respectively.
Regarding the BoW dictionaries used in our experiments, the one for dense SIFT was 200-dimensional, while another for RGB values was 150-dimensional. The two dictionaries were learned using natural scene images from PASCAL VOC 2012.
With respect to the concatenated color histograms, 16 bins for RGB, HSV, LAB, and YCbCr were used in each channel, and thus a 192-dimensional color histogram was obtained for each node.

\subsection{\textbf{ICE vs. Separate Encoding}}
In this experiment part, we compare our ICE against the conventional separate encoding.
The quantitative experimental results for five SegTrack videos are shown in Table \ref{table:ICE}.

Here, we extracted the optical flow cues based on \cite{papazoglou2013fast} and appearance saliency based on \cite{yang2013saliency}, and then combined them in the final stage as a likelihood map denoted as separate encoding (SE). We compare SE with ICE using the same energy model and the same parameters. In the simplified energy minimization model, the pre- and post-processing procedures are discarded, and only ICE and SE maps are respectively employed to build unary potentials. With respect to pairwise potentials, it is impossible to measure the pairwise cost of two adjacent nodes using just the two likelihood maps. For simplicity, the absolute difference between two adjacent nodes is used as their pairwise potential.


As Table \ref{table:ICE} shows, ICE obviously outperforms SE under the same condition.
The improvement is mainly caused by exploiting the homologous properties of multimodal cues for the same target, and use them throughout the whole processing framework.
%

\begin{table}[tbp]  
\centering
\begin{tabular}{c|ccccc|c}  
\hline
Methods                 &birdfall &cheetah &girl &monkeydog  &parachute  &Avg.   \\ \hline  
SE                        &447 &1626 &4217  &1576    &627     &1450 \\  
ICE                      &379  &1381  &2432   &951   &396   &942 \\         \hline
\end{tabular}
\caption{Comparison of interactively constrained encoding (ICE) and separate encoding (SE) on SegTrack. The metric is average per-frame pixel error.} \label{table:ICE}
\end{table}

\subsection{\textbf{Results and Analysis on SegTrack}}
\begin{figure*}[htbp]  
\centering
  \hspace{-0.1in} \vspace{-0.13in}
  \subfigure{ \includegraphics[width=0.21\textwidth, height=1.0in]{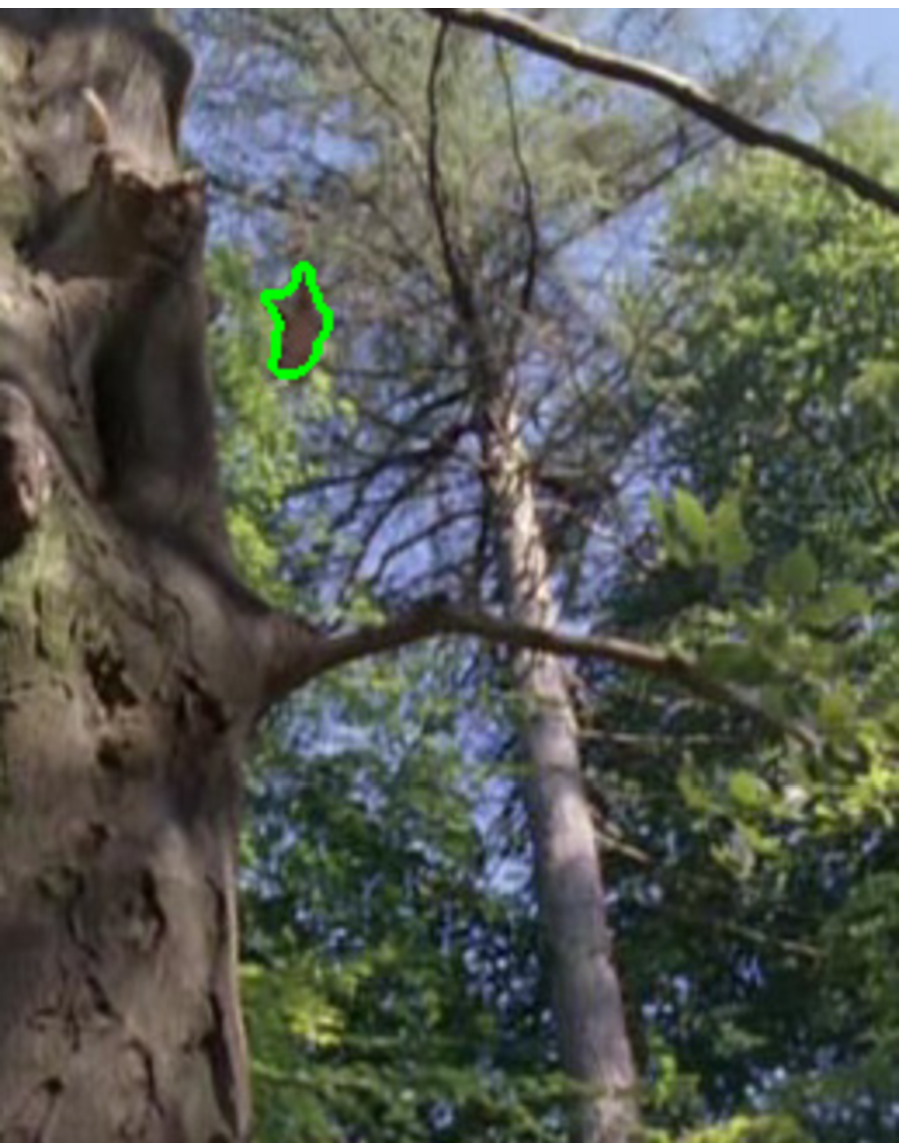} }
  \hspace{-0.1in}
  \subfigure{ \includegraphics[width=0.21\textwidth, height=1.0in]{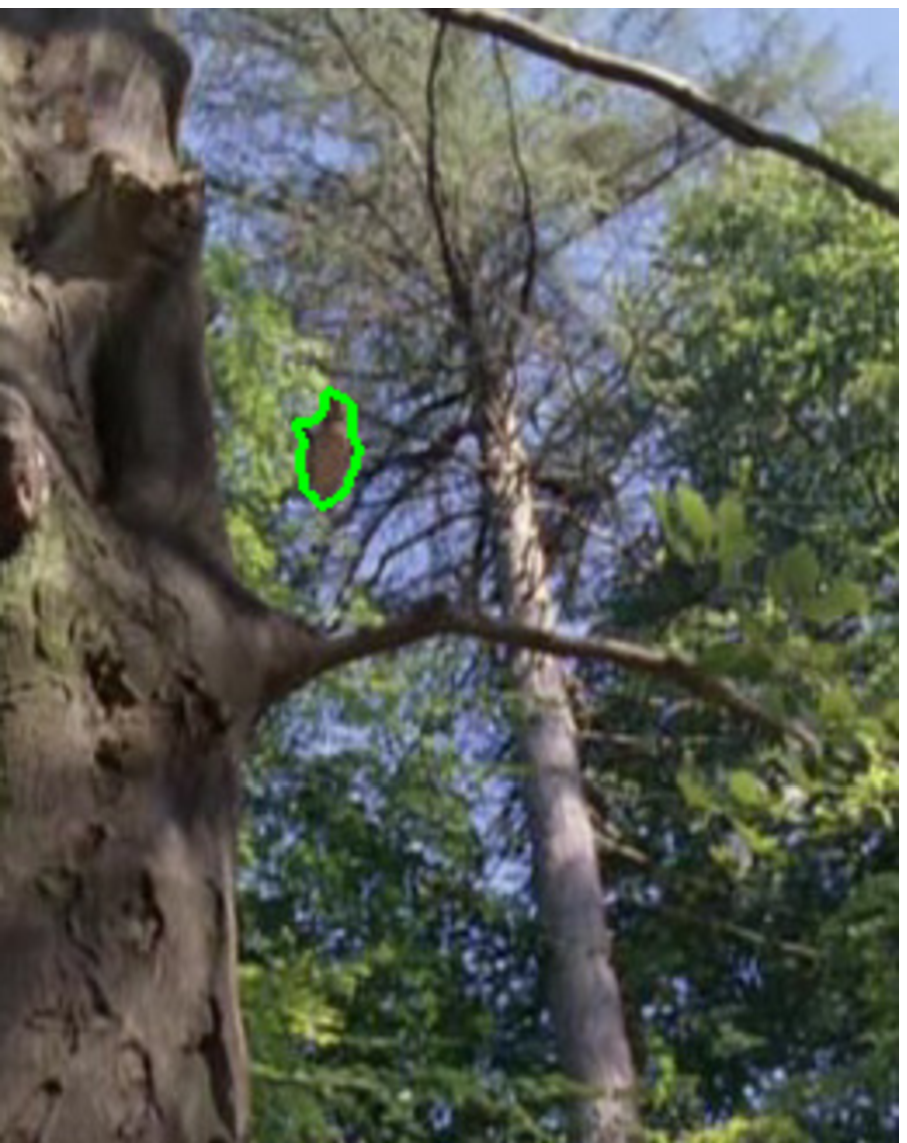} }
  \hspace{-0.1in}
  \subfigure{ \includegraphics[width=0.21\textwidth, height=1.0in]{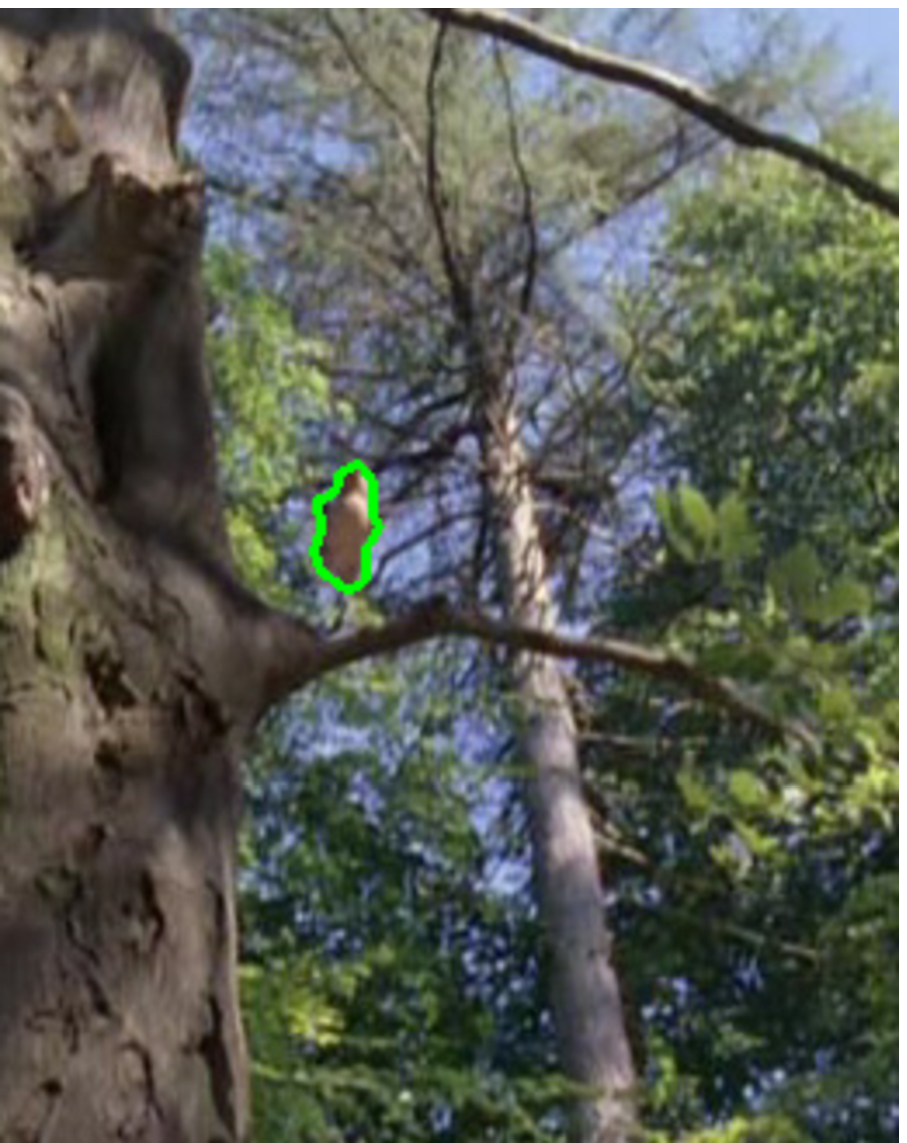} }
  \hspace{-0.1in}
  \subfigure{ \includegraphics[width=0.21\textwidth, height=1.0in]{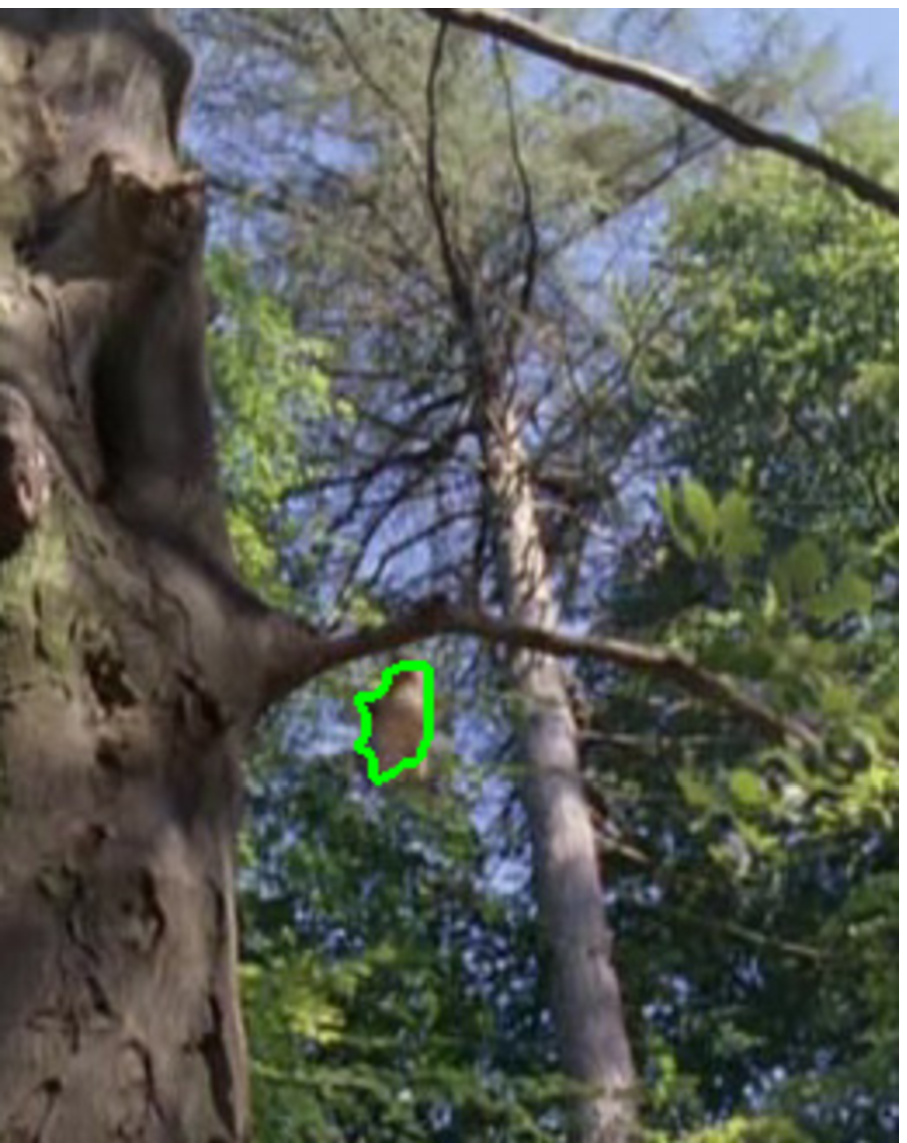} }

  \hspace{-0.1in} \vspace{-0.13in}
  \subfigure{ \includegraphics[width=0.21\textwidth]{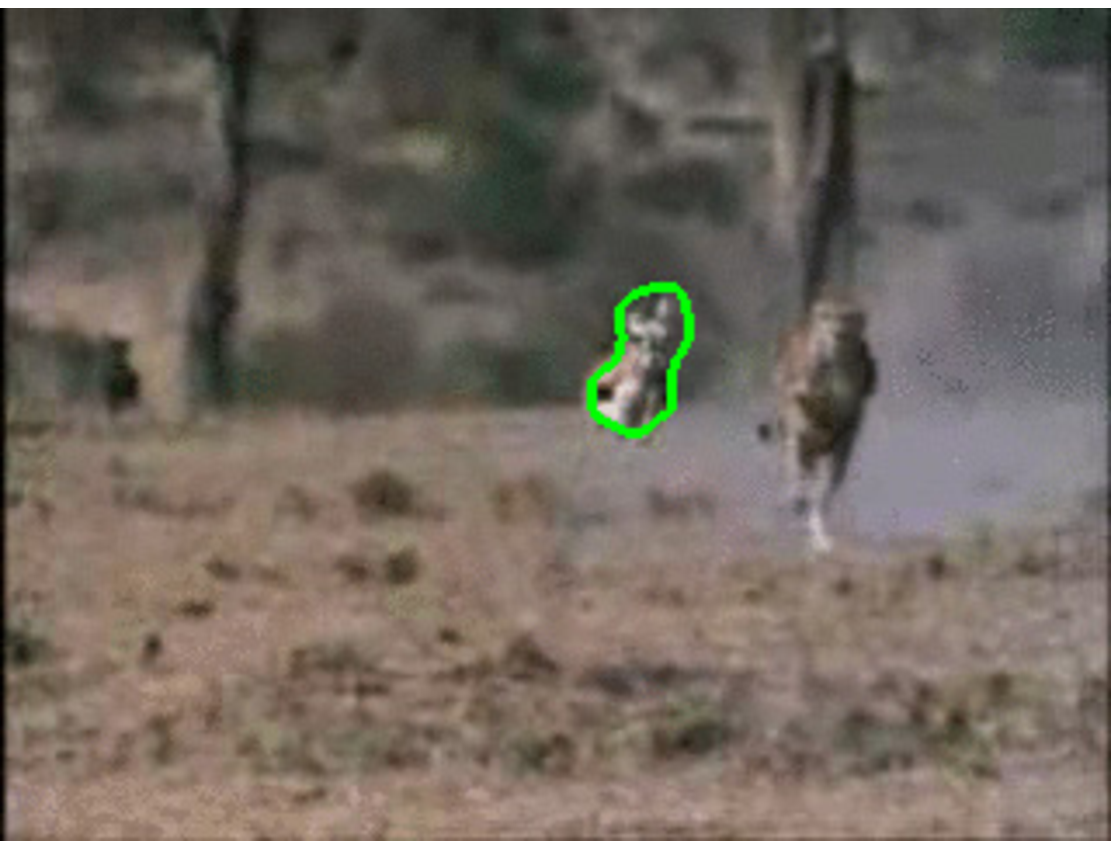} }
  \hspace{-0.1in}
  \subfigure{ \includegraphics[width=0.21\textwidth]{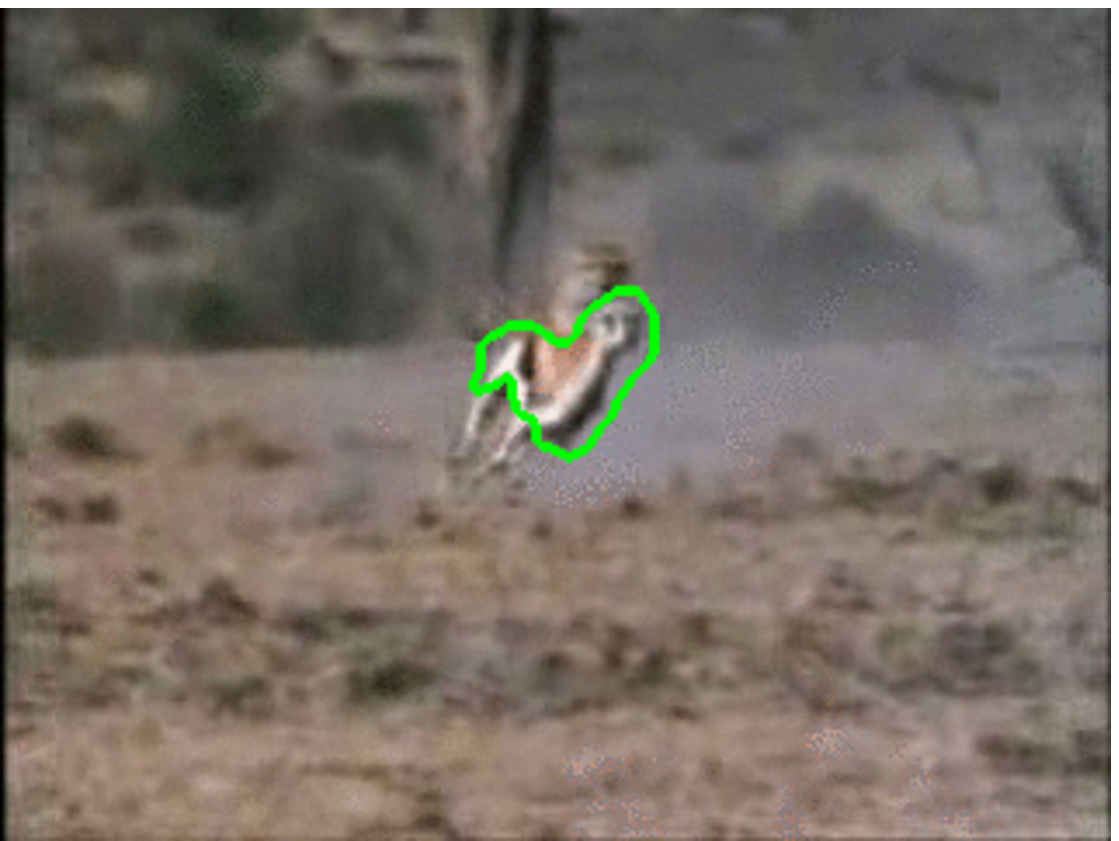} }
  \hspace{-0.1in}
  \subfigure{ \includegraphics[width=0.21\textwidth]{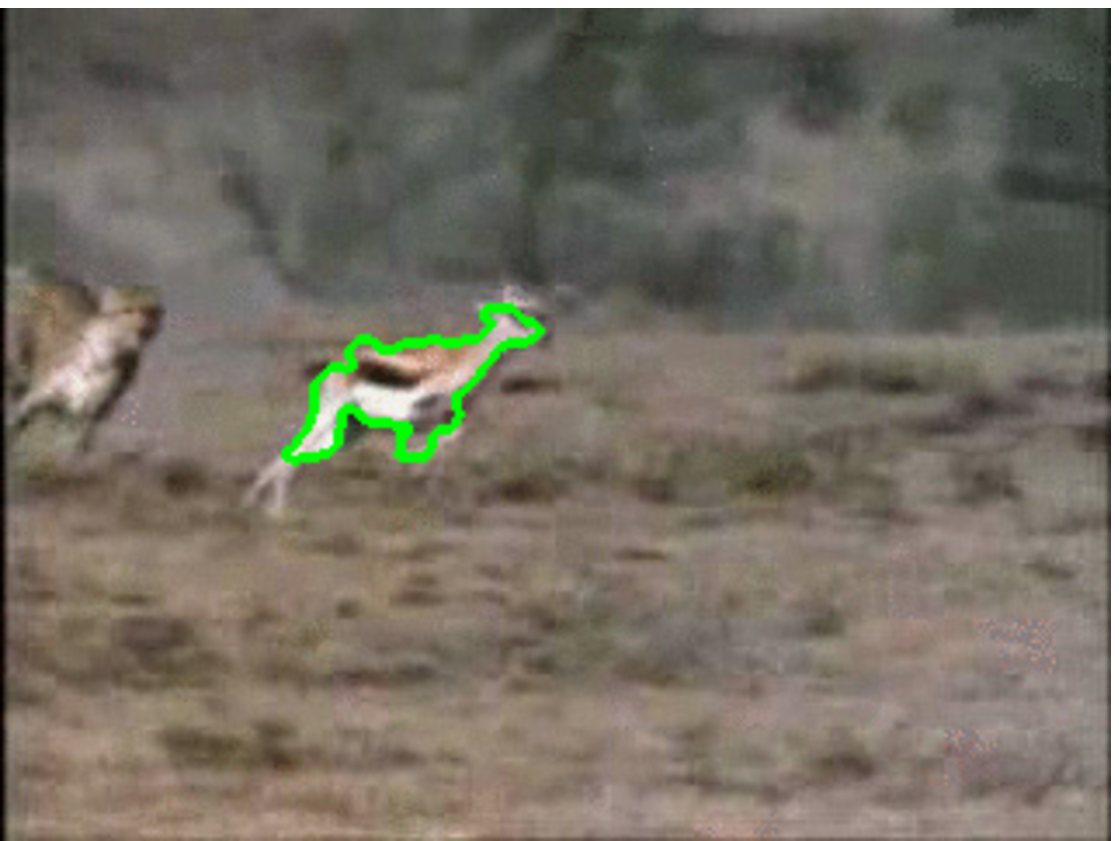} }
  \hspace{-0.1in}
  \subfigure{ \includegraphics[width=0.21\textwidth]{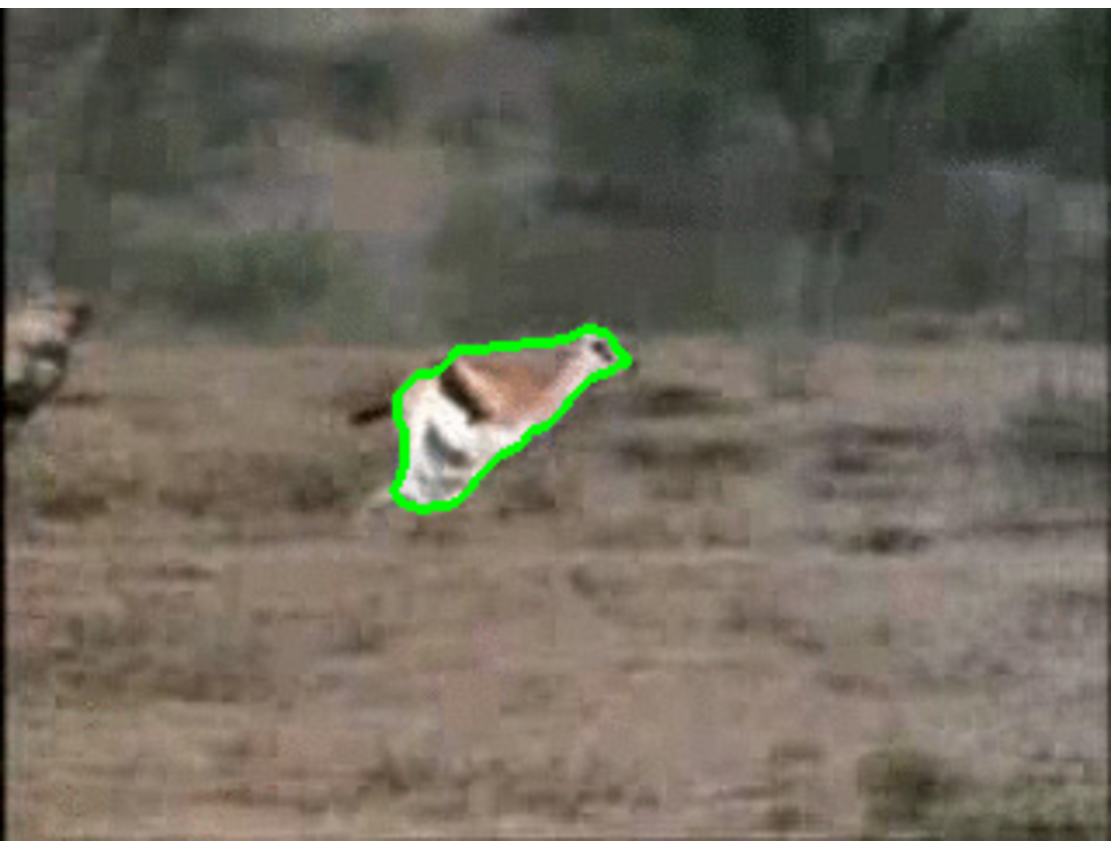} }

  \hspace{-0.1in} \vspace{-0.13in}
  \subfigure{ \includegraphics[width=0.21\textwidth]{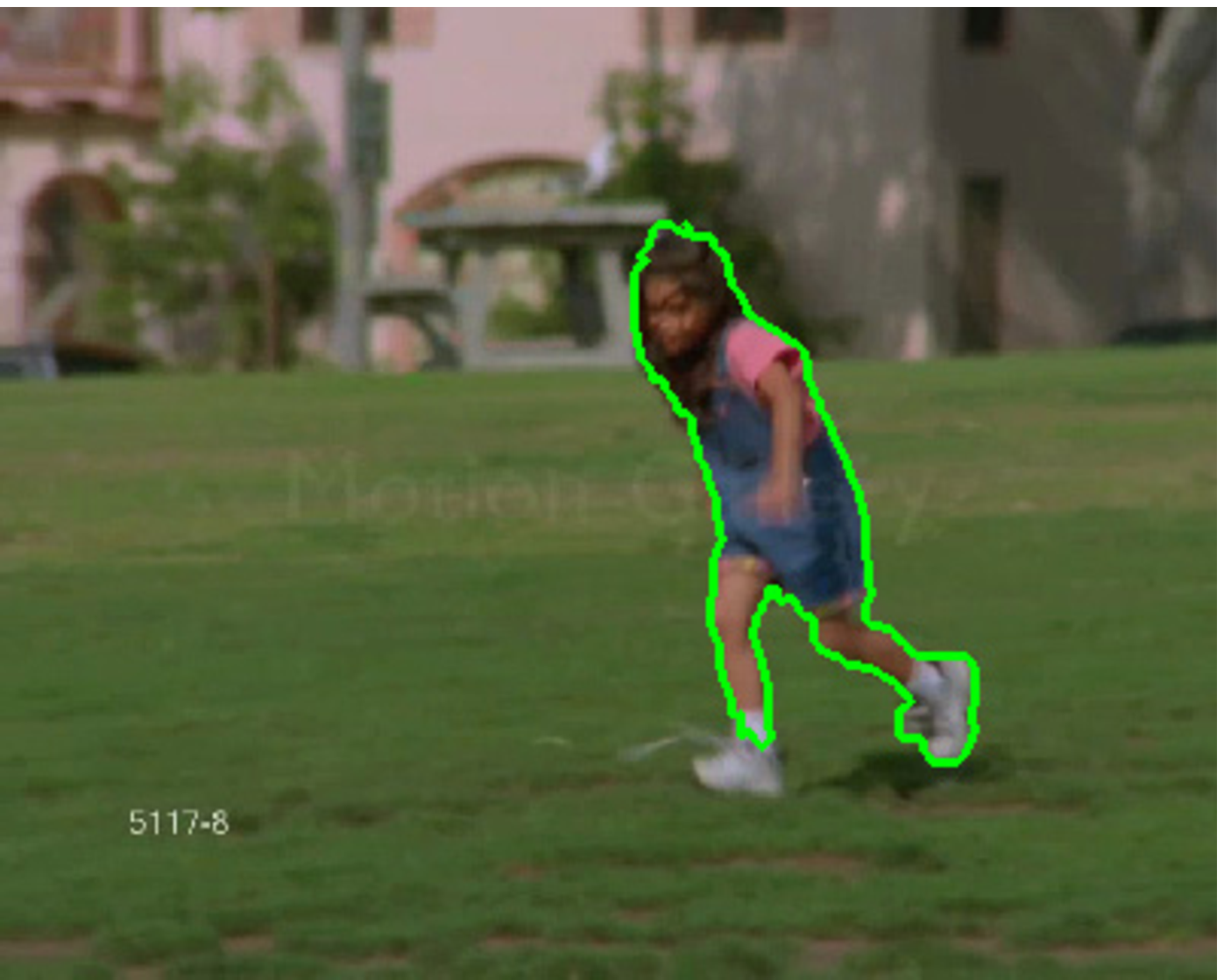} }
  \hspace{-0.1in}
  \subfigure{ \includegraphics[width=0.21\textwidth]{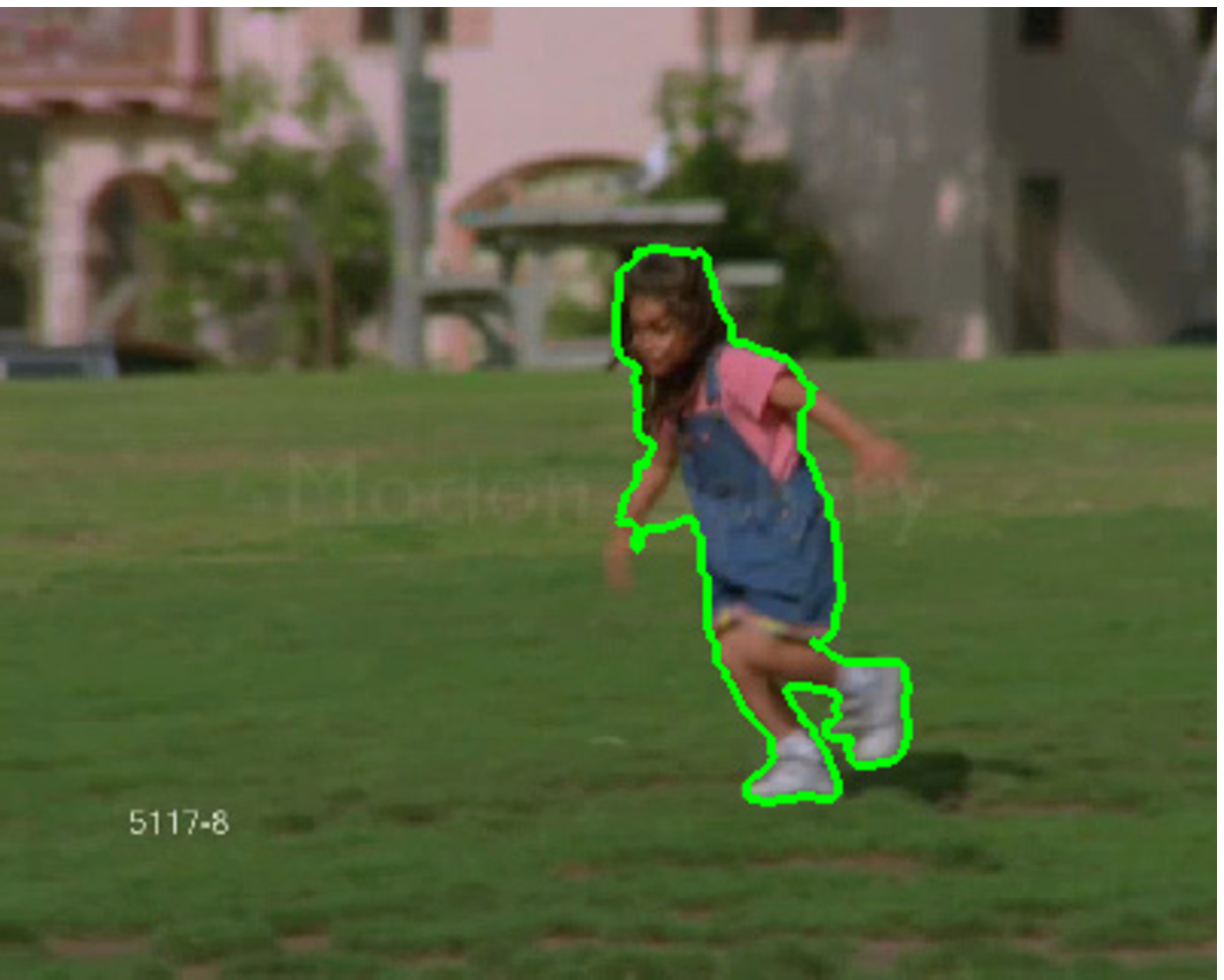} }
  \hspace{-0.1in}
  \subfigure{ \includegraphics[width=0.21\textwidth]{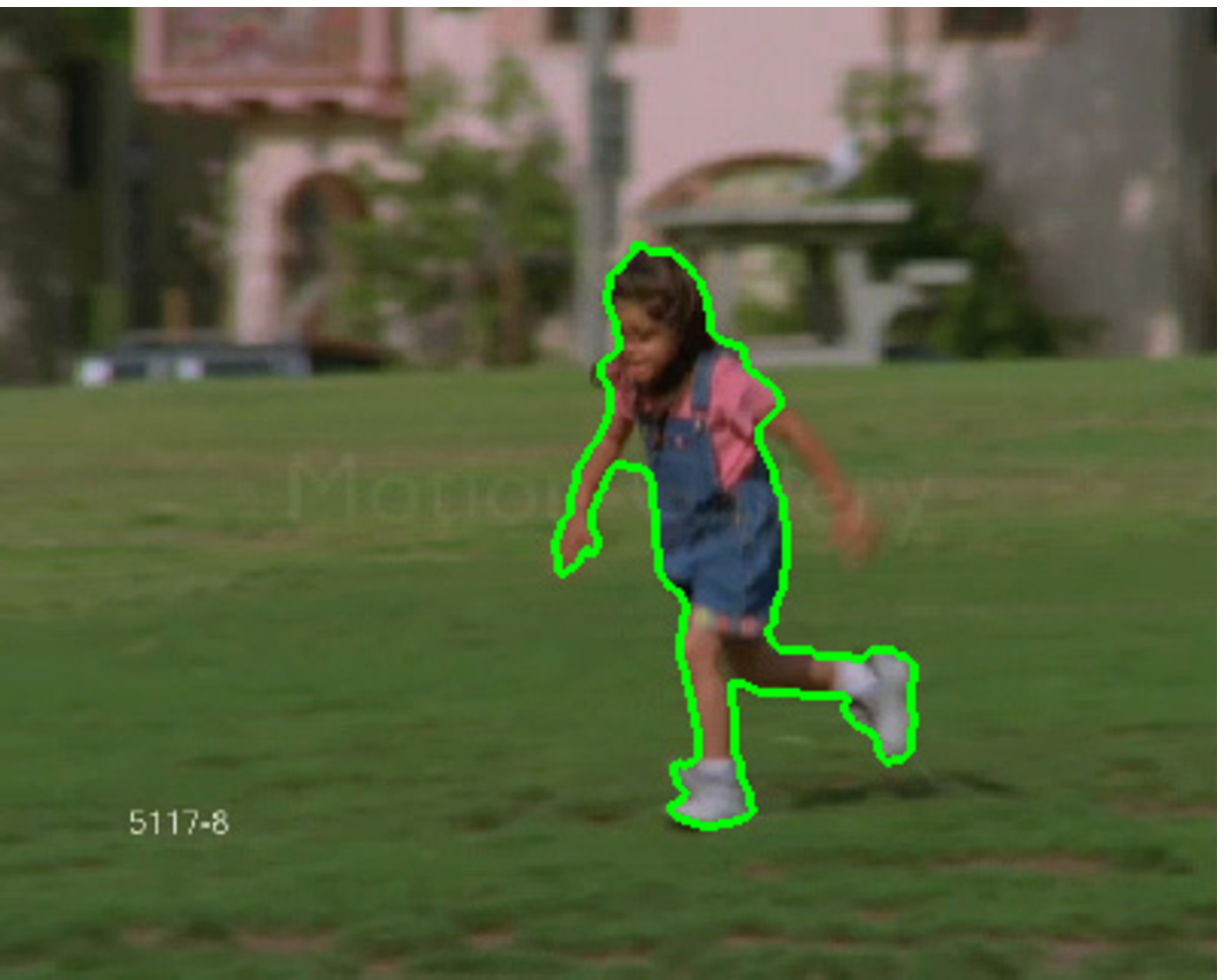} }
  \hspace{-0.1in}
  \subfigure{ \includegraphics[width=0.21\textwidth]{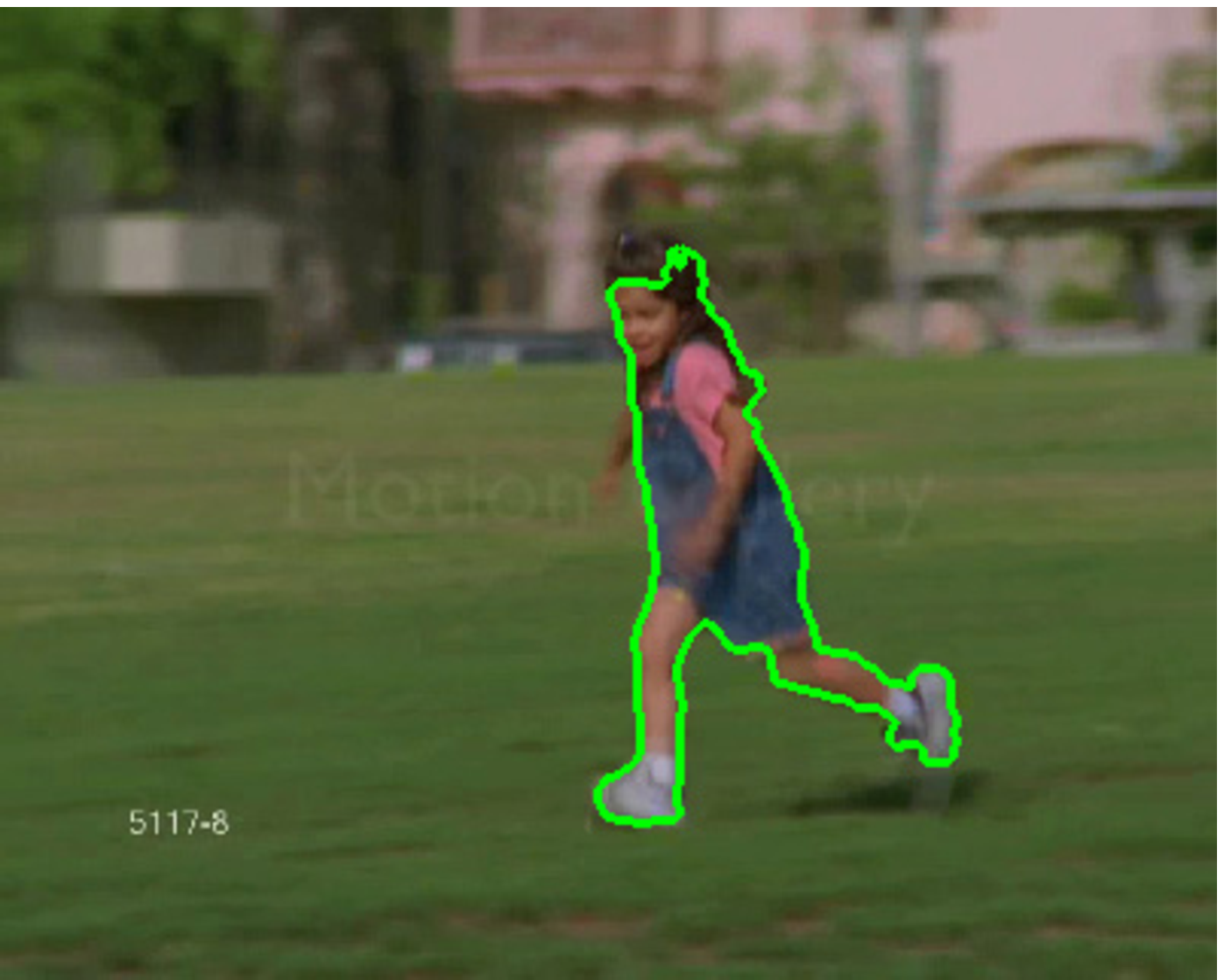} }

  \hspace{-0.1in} \vspace{-0.13in}
  \subfigure{ \includegraphics[width=0.21\textwidth]{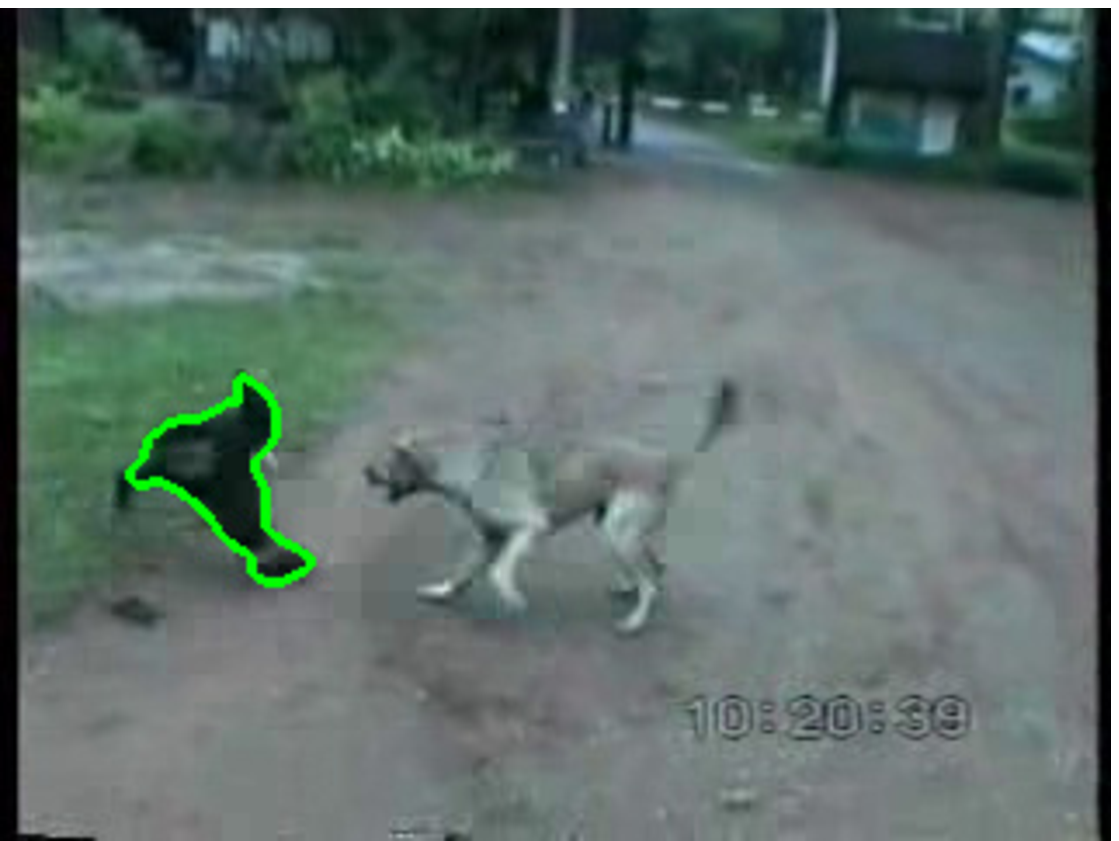} }
  \hspace{-0.1in}
  \subfigure{ \includegraphics[width=0.21\textwidth]{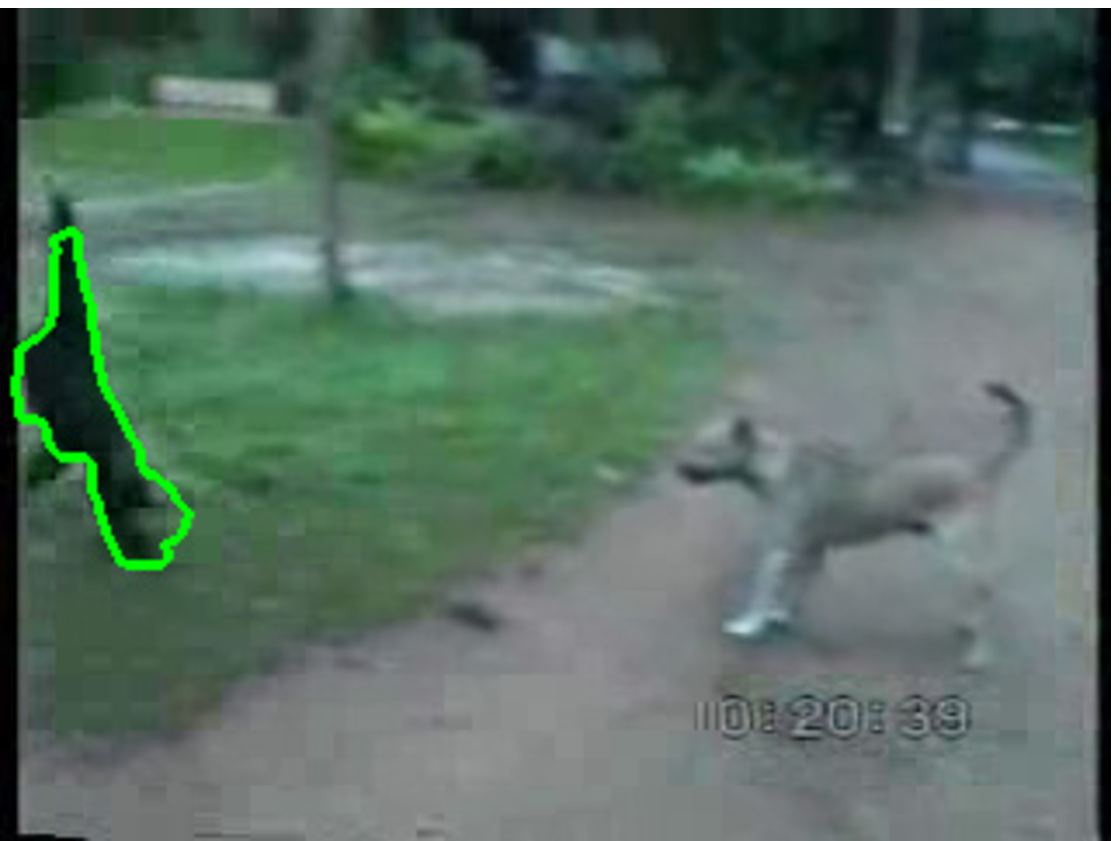} }
  \hspace{-0.1in}
  \subfigure{ \includegraphics[width=0.21\textwidth]{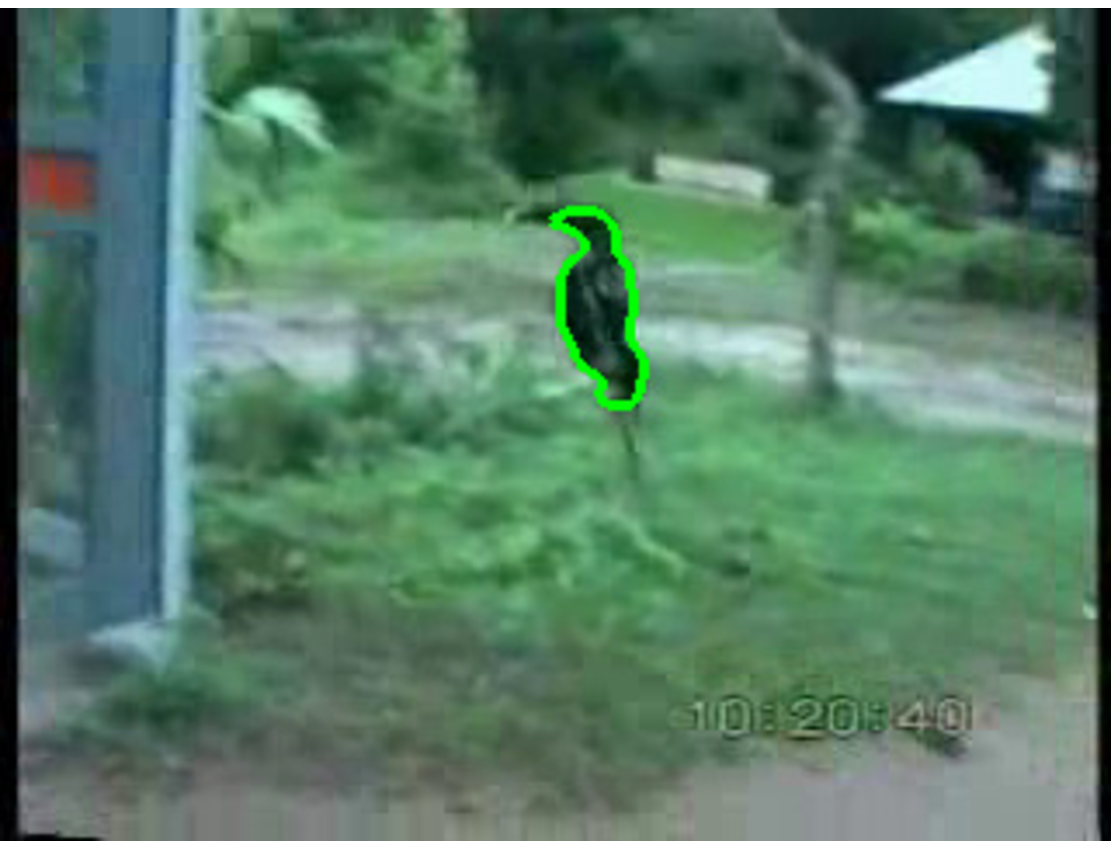} }
  \hspace{-0.1in}
  \subfigure{ \includegraphics[width=0.21\textwidth]{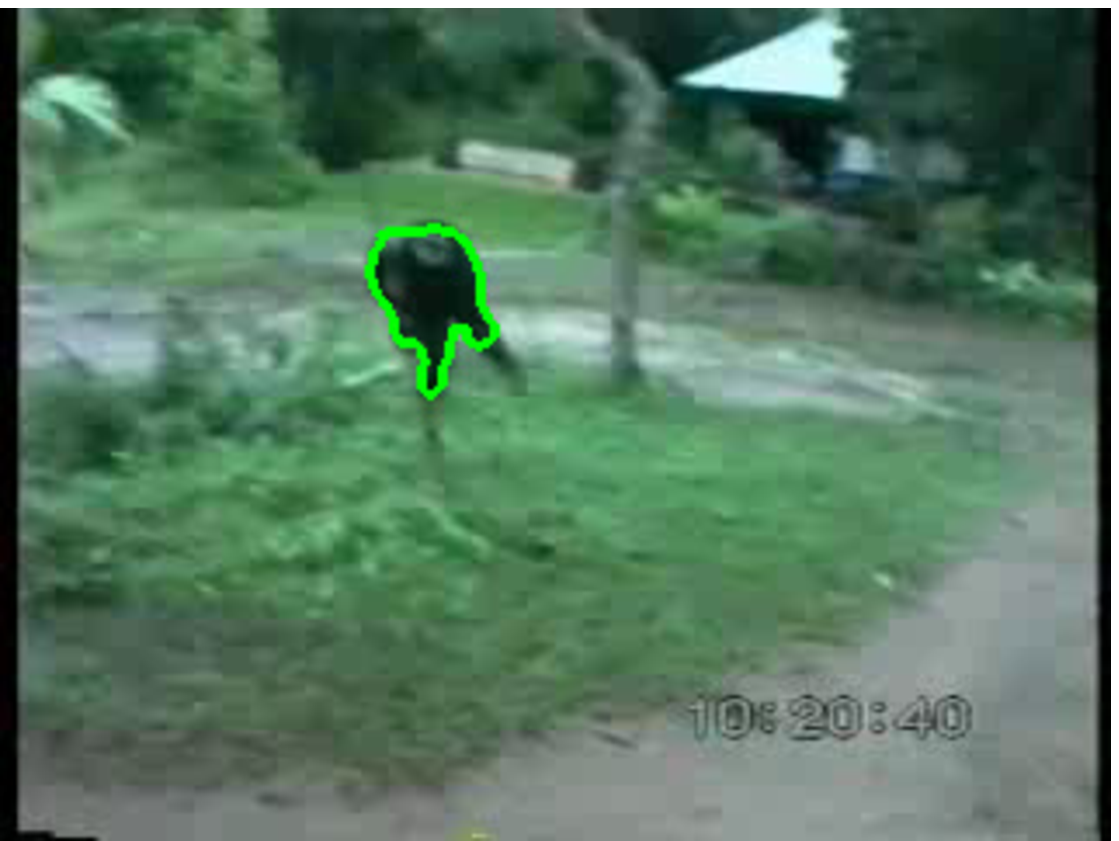} }

  \hspace{-0.1in} \vspace{-0.13in}
  \subfigure{ \includegraphics[width=0.21\textwidth]{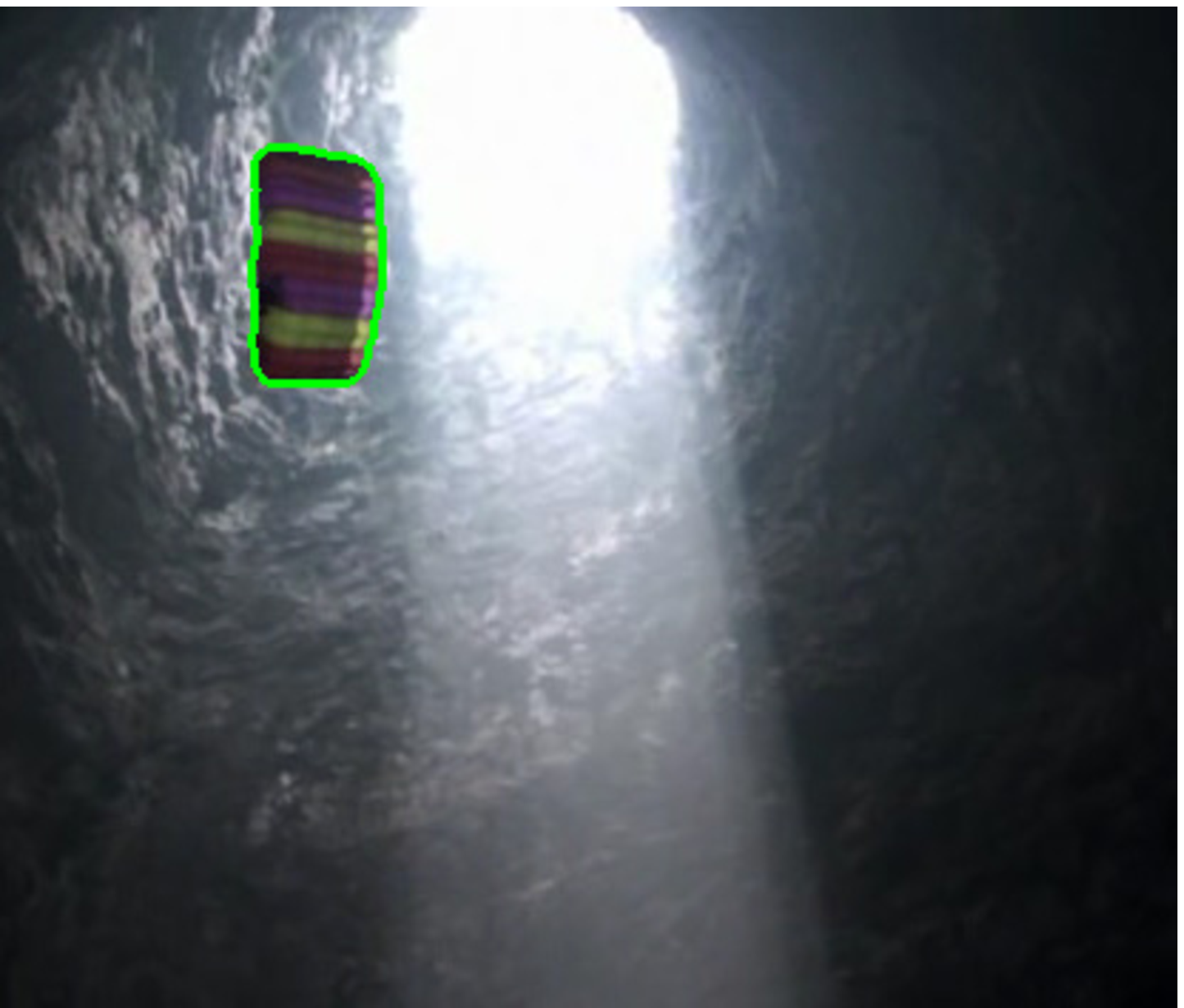} }
  \hspace{-0.1in}
  \subfigure{ \includegraphics[width=0.21\textwidth]{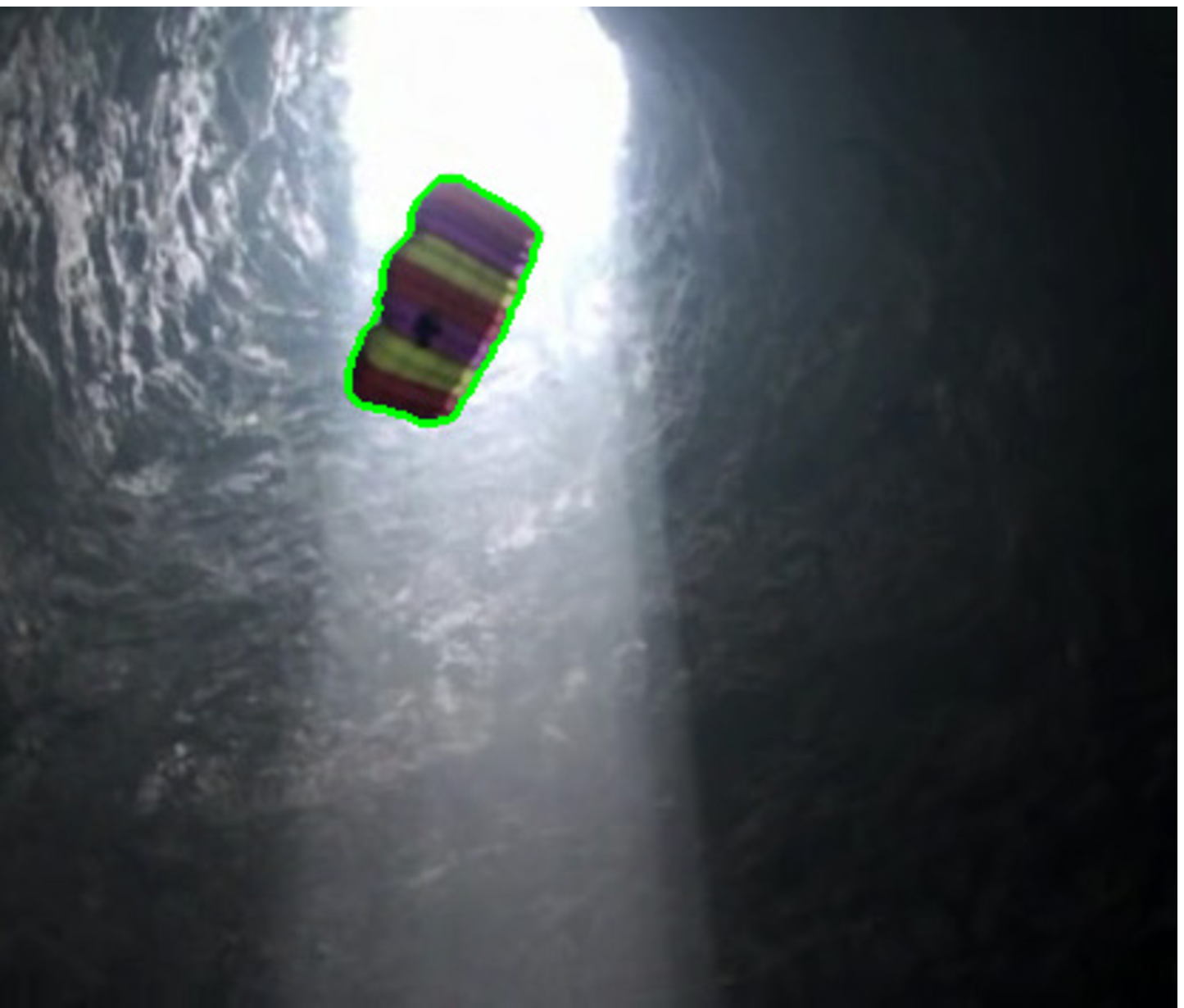} }
  \hspace{-0.1in}
  \subfigure{ \includegraphics[width=0.21\textwidth]{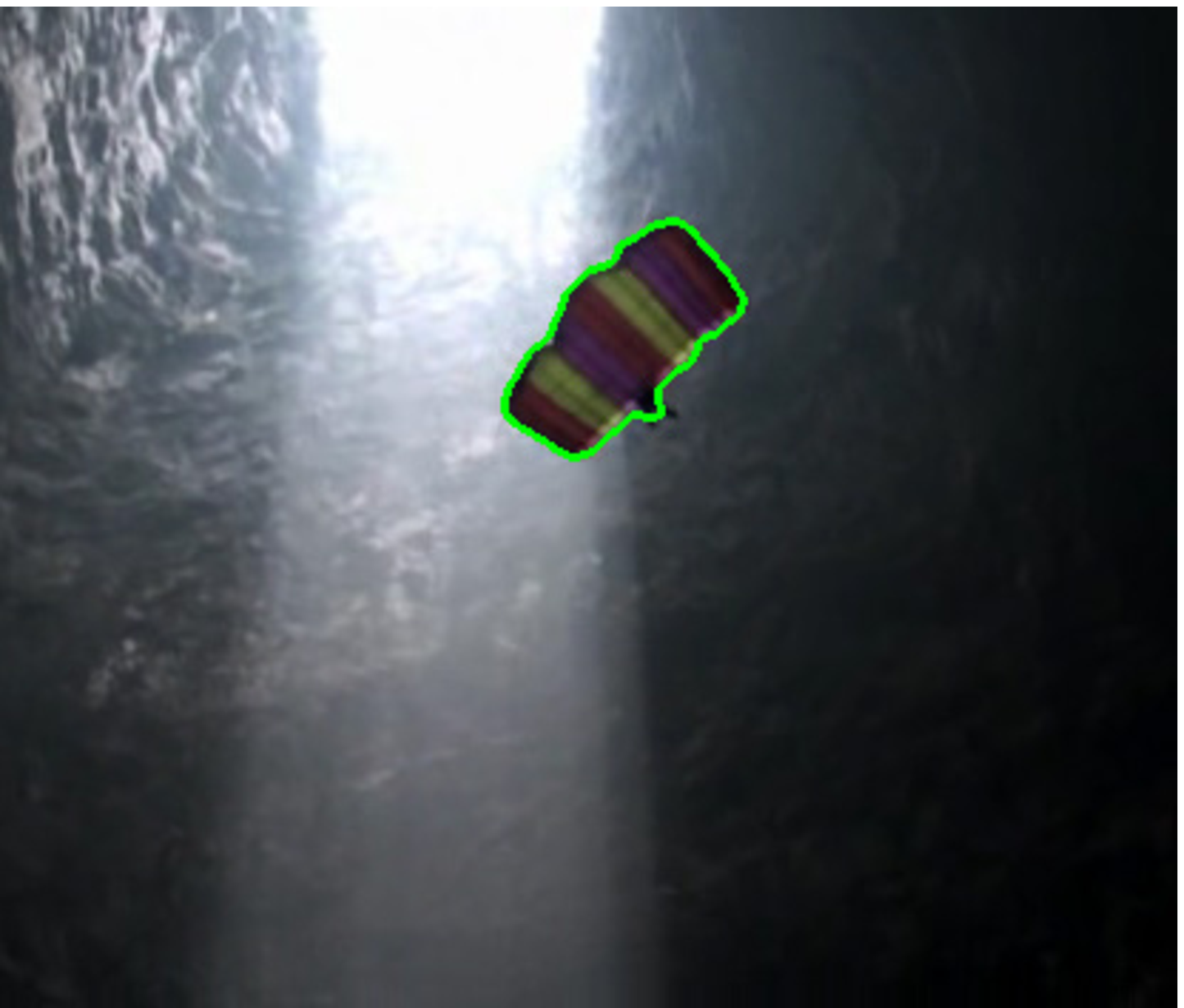} }
  \hspace{-0.1in}
  \subfigure{ \includegraphics[width=0.21\textwidth]{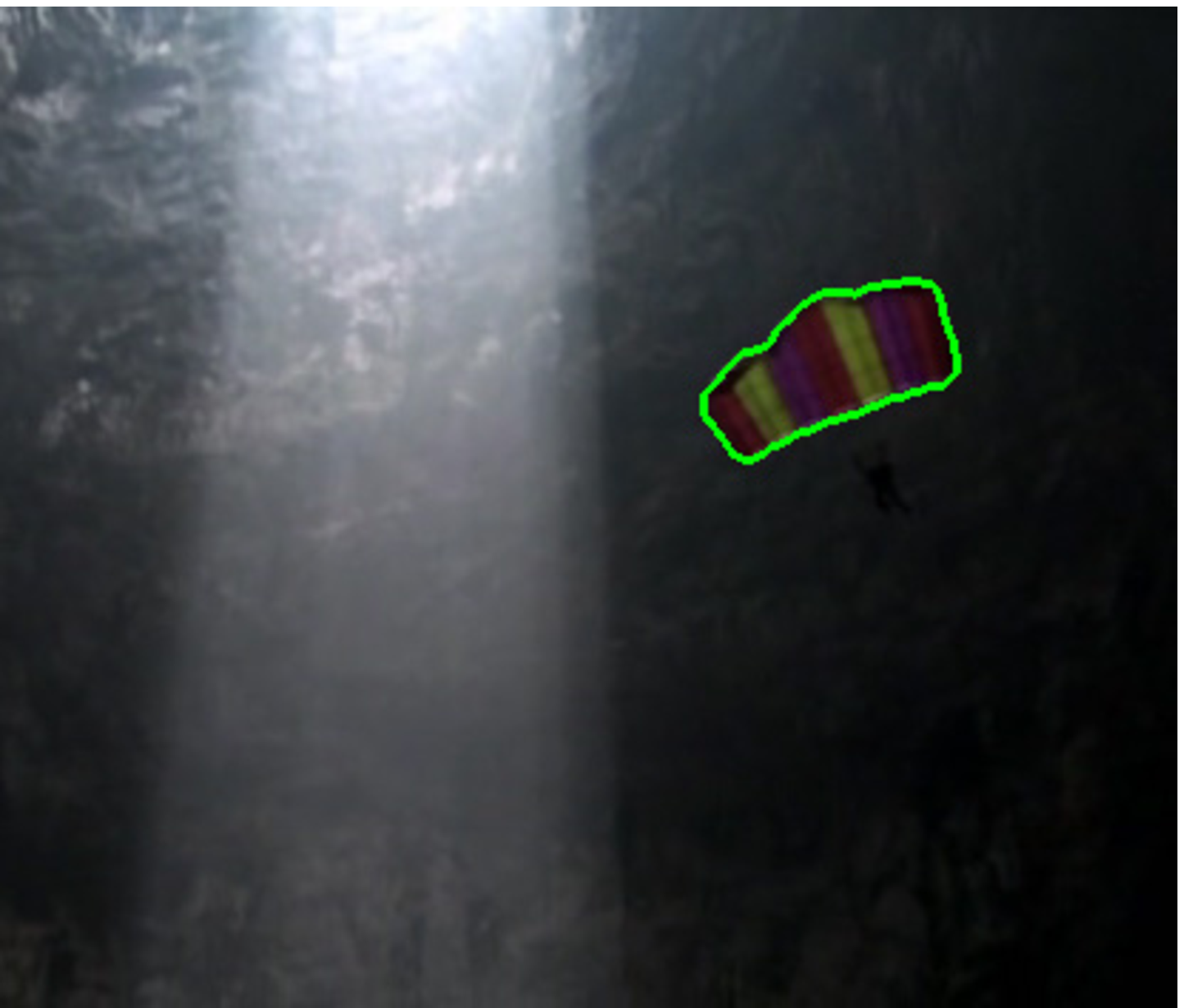} }

\caption{Illustration of object segmentation results by our method on SegTrack. Results from top to bottom rows are respectively from videos: \textbf{\emph{birdfall, cheetah, girl, monkeydog}}, and \textbf{\emph{parachute}}. Segmented objects are delimitated by green curves. Best view in color.}
\label{figBird}
\end{figure*}

Comprehensive comparisons of object segmentation results on SegTrack are demonstrated in Table \ref{table:segtrack}, which includes our approach and nine state-of-the-art methods. In addition to the average per-frame pixel error for each video and the whole video set, we enumerate what the basic unit is when optimizing the model.
Similar to \cite{papazoglou2013fast}, we process objects at the superpixel level, whereas \cite{zhang2013video,wang2015saliency} and other approaches operate at the pixel level.
\cite{tsai2010motion} and \cite{chockalingam2009adaptive} are conducted in a supervised manner that requires manual annotation in the first frame, whereas others are not.
Our approach achieves a competitive result, especially comparing with the method presented in \cite{papazoglou2013fast}, which also uses superpixels as basic units.
Although sacrificing some accuracy compared with pixel-by-pixel optimization \cite{zhang2013video,wang2015saliency}, our approach is very feasible and efficient, as the number of units in the model is reduced by two orders of magnitude.
To investigate related algorithms under a normalized metric, we also calculate the average precision using our newly defined metric in Eq.\eqref{avgPrecision}, which measures the percentage of correctly labeled pixels. As Table \ref{table:segprecision} shows, our approach obtains an average precision of 99.4\%. Under this normalized metric, we can see that performances of most approaches are very close and eligible.
The qualitative results for the five sequences are also given in Figure \ref{figBird}.

\setlength{\tabcolsep}{5pt} 

\begin{table*}[htbp]

\centering
\begin{tabular}{l|cccccccccc}  
\hline
Video &\scriptsize{Ours} &\scriptsize{Papazoglou's\cite{papazoglou2013fast}} &\scriptsize{Wang's\cite{wang2015saliency}} &\scriptsize{Varas's\cite{varas2014region}} &\scriptsize{Ochs's\cite{ochs2012higher}} &\scriptsize{Zhang's\cite{zhang2013video}} &\scriptsize{Lee's\cite{lee2011key}} &\scriptsize{Brox's\cite{brox2010object}} &\scriptsize{Tsai's\cite{tsai2010motion}} &\scriptsize{Chockalingam's\cite{chockalingam2009adaptive}}  \\ \hline  
\scriptsize{birdfall}  &219  &217  & 209   &243   &468   &155  &288  &468  &252  &454      \\         
\scriptsize{cheetah}   &834  &890  & 796   &391   &1175  &633  &905  &1968 &1142 &1217    \\        
\scriptsize{girl}      &1512 &3859 & 1040  &1935  &5683  &1488 &1785 &7595 &1304 &1755    \\
\scriptsize{monkeydog} &536  &284  & 562   &497   &1434  &365  &521  &1434 &563  &683     \\
\scriptsize{parachute} &353  &855  & 207   &187   &1595  &220  &201  &1113 &235  &502    \\ \hline
\scriptsize{Avg.}       &587  &877  & 503   &515   &1736  &452  &592  &1926 &594  &791     \\ \hline
\scriptsize{Basic Unit} &sp  &sp    &p  &p &p &p &p &p &p   &p   \\  \hline
\scriptsize{Supervised?}  &$\times$  & $\times$    & $\times$   & $\times$   & $\times$   &$\times$   & $\times$   &$\times$  & $\surd$  & $\surd$    \\ \hline

\end{tabular}

\caption{Average per-frame pixel error on SegTrack} \label{table:segtrack}
\end{table*}

\begin{table*}[htbp]
\centering

\begin{tabular}{l|cccccccccc}  
\hline
Video &\scriptsize{Ours} &\scriptsize{Papazoglou's\cite{papazoglou2013fast}} &\scriptsize{Wang's\cite{wang2015saliency}} &\scriptsize{Varas's\cite{varas2014region}} &\scriptsize{Ochs's\cite{ochs2012higher}} &\scriptsize{Zhang's\cite{zhang2013video}} &\scriptsize{Lee's\cite{lee2011key}} &\scriptsize{Brox's\cite{brox2010object}} &\scriptsize{Tsai's\cite{tsai2010motion}} &\scriptsize{Chockalingam's\cite{chockalingam2009adaptive}}  \\ \hline  
\scriptsize{birdfall}  &99.7  &99.7  & 99.8   &99.7   &99.4  &99.8  &99.7  &99.4  &99.7  &99.5      \\         
\scriptsize{cheetah}   &98.9  &98.8  & 99.0   &99.5   &98.5  &99.2  &98.8  &97.4 &98.5 &98.4    \\        
\scriptsize{girl}      &98.8 &97.0   &99.2    &98.5   &95.6  &98.8  &98.6  &94.1 &99.0 &98.6    \\
\scriptsize{monkeydog} &99.3  &99.6  & 99.3   &99.4   &98.1  &99.5  &99.3  &98.1 &99.3  &99.1     \\
\scriptsize{parachute} &99.8 &99.4  & 99.9   &99.9    &98.9  &99.8  &99.9  &99.2 &99.8  &99.7    \\ \hline
\scriptsize{Avg.}       &99.4  &99.1  & 99.5   &99.5   &98.3  &99.6  &99.4  &98.1 &99.4  &99.2     \\ \hline
\end{tabular}
\caption{Average precision $\left ( \% \right )$ on SegTrack} \label{table:segprecision}
\end{table*}

\begin{figure*}[tbp]
\hspace{-0.8in}
\includegraphics[width=0.7\textwidth]{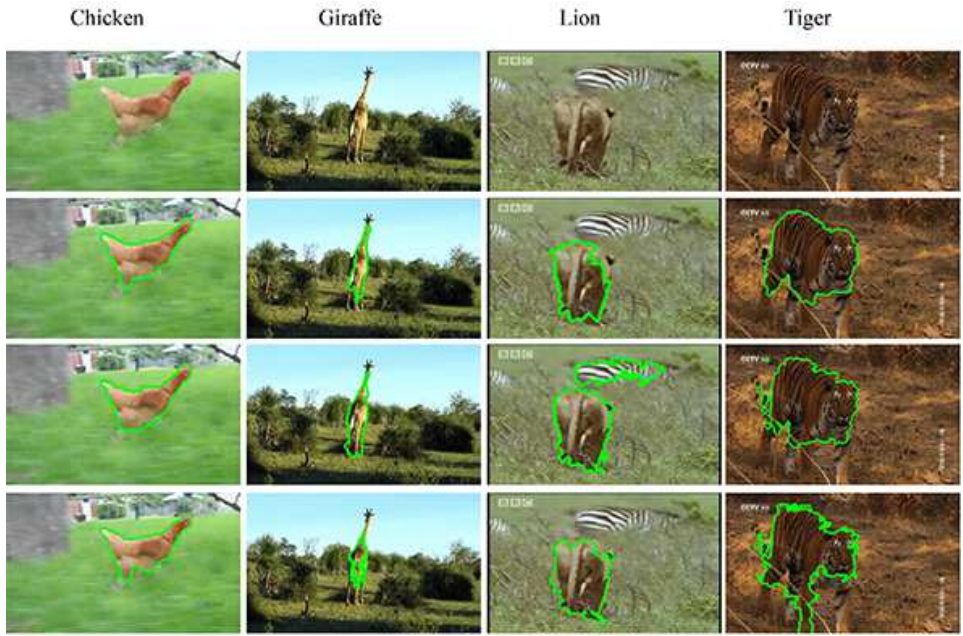}
 \centering
\caption{Comparison of different methods on MOViCS. Four rows from top to bottom show \textbf{\emph{original frames, our results, segmentations from \cite{papazoglou2013fast}}}, and \textbf{\emph{results from \cite{chiu2013multi}}}, respectively. Best view in color.}
\label{figMOViCS}
\end{figure*}

\begin{figure*}[htbp]
\centering
\hspace{-1.1in}
   \subfigure[Yuna Kim]{ \includegraphics[width=0.8\textwidth]{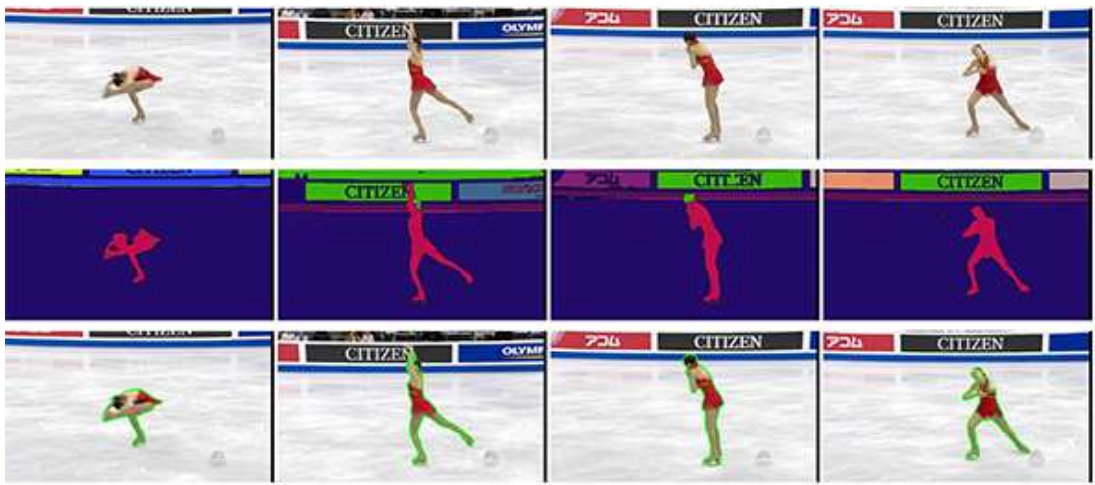} }

\hspace{-1.1in}
  \subfigure[waterski]{ \includegraphics[width=0.8\textwidth]{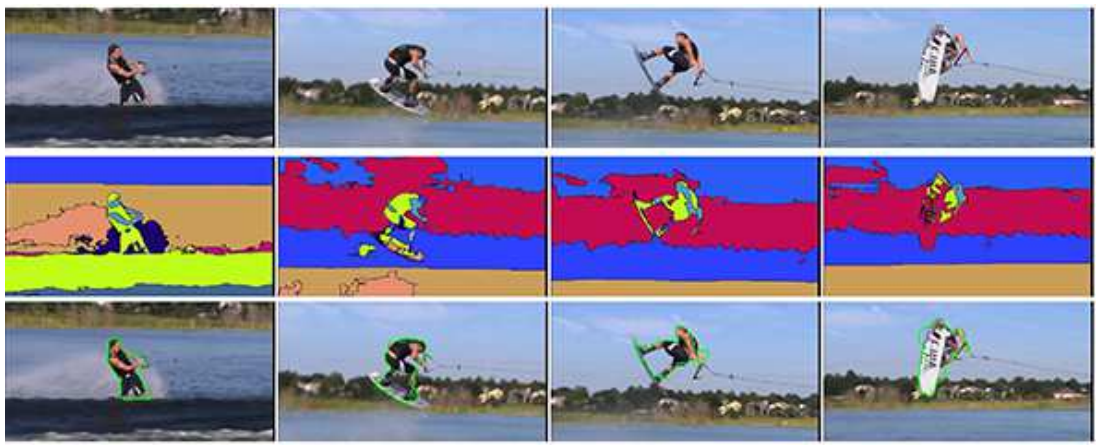} }
\caption{Comparisons between our method and \cite{tsai2010motion} on GaTech. In each subfigure, three rows from top to bottom, show \textbf{\emph{original frames, segmentations from \cite{tsai2010motion}}}, and \textbf{\emph{our results}}, respectively. Best view in color.}
\label{figGaTech}
\end{figure*}

\subsection{\textbf{Results and Analysis on MOViCS}} \label{sec:resultsMovics}
There are four sets in MOViCS, each of which contains 2 to 4 relevant videos. As MOViCS \cite{chiu2013multi} is originally collected for object co-segmentation, the identical object occurs repeatedly in relevant videos. However, in some videos, target objects are always nearly static, which is beyond the scope of this paper, which aims to segment moving objects. Thus, we only use one video sequence containing moving targets for each set, and four sequences in total are eventually used as test data.
We compare our method against \cite{papazoglou2013fast} and \cite{chiu2013multi}, running the codes provided by the authors on MOViCS. The parameter values are also the same as those set by the codes.
Table \ref{table:MOViCS} demonstrates the average per-frame pixel errors of different methods on MOViCS, and Table \ref{table:precisionMOViCS} shows the average precision of the three methods. Additionally, visual comparisons of segmentation are also depicted in Figure \ref{figMOViCS}.

\setlength{\tabcolsep}{4.5pt} 

\begin{table}[tb]  
\centering
\begin{tabular}{l|cccc|c|c}  
\hline
Methods                 &Chicken &Giraffe &Lion &Tiger  &Avg.    &Supervised?\\ \hline  
Ours                      &1437  &2153  &2605   &14960    &5289   & $\times$ \\         
Papazoglou's\cite{papazoglou2013fast} &1762  &2225  &2670  &17480   &6034    & $\times$ \\        
Chiu's\cite{chiu2013multi}      &1271 &1352 &3250  &6158    &3008      & $\surd$ \\ \hline

\end{tabular}
\caption{Average per-frame pixel error on MOViCS.} \label{table:MOViCS}
\end{table}

Note that \cite{chiu2013multi} is weakly supervised and requires relevant videos containing the identical targets as input. Hence, during the experiments, we still fed all of the relevant videos in MOViCS to the model in \cite{chiu2013multi}, but only selected the four videos that were both used by us and by \cite{papazoglou2013fast} to report the performance.
Additionally, the approach developed by \cite{chiu2013multi} outputs a set of segments containing both background and foreground for each frame, and those segments do not specify which one belongs to the foreground moving target. Hence, we compare each segment to the ground-truth foreground one by one and adopt the segment with the highest accuracy as the foreground. That means that unlike the settings in \cite{papazoglou2013fast} and in our study, the performance of \cite{chiu2013multi} in Table \ref{table:MOViCS} is actually aided by both the ground truth and extra relevant videos.

\begin{table}[tb]
\centering
\begin{tabular}{l|cccc|c}  
\hline
Methods                 &Chicken &Giraffe &Lion &Tiger  &Avg.    \\ \hline  
Ours                      &98.9  &98.3  &96.6   &93.5    &96.3   \\         
Papazoglou's\cite{papazoglou2013fast} &98.6  &98.3  &96.5  &92.4   &95.7    \\        
Chiu's\cite{chiu2013multi}      &99.0 &99.0  &95.8 &97.3    &97.9     \\ \hline

\end{tabular}
\caption{Average precision $\left ( \% \right )$ on MOViCS.} \label{table:precisionMOViCS}
\end{table}


\subsection{\textbf{Results and Analysis on GaTech}}
Given that pixel-level ground truth is not offered in GaTech \cite{tsai2010motion}, we qualitatively compare our method with \cite{tsai2010motion} on this dataset. The visual comparisons are shown in Figure \ref{figGaTech}, where our segmentations are delimitated by green curves.

Different from our method that is unsupervised, \cite{tsai2010motion} is a supervised approach that usually works well under uniform backgrounds (e.g., Yuna Kim), but can introduce a host of object fragments due to scene clutter (e.g., waterski). This is mainly because \cite{tsai2010motion} does not integrate object-level cues, whereas our approach provides more insights into object-oriented information, such as the saliency and accumulated proposal maps in our ICE.


\section{Conclusion}
In this paper, we present an unsupervised approach for moving object segmentation in unconstrained videos. The interactively constrained encoding (ICE) is proposed to exploit the homologous properties of multimodal cues for the same object. Due to preserving interactive restrictions throughout both the initialization and refinement stages, our approach can well perceive and refine moving targets in variant environments, even under a failure of appearance saliency or optical flow. We also partially tackle the inter-occlusion problem through a conservative proposal-wise maximum bipartite graph matching.
Furthermore, the lightweight superpixel-level graph optimization is developed to reduce the computation complexity.
Future work involves incorporating more geometric patterns and perceptual information into the approach, and using multiple adjacent frames to generate temporal ICE.


\section*{Acknowledgement}
This work was supported by the NSFC (No. U1611461, 61573387, 61672544) and the Guangdong Program (No. 2015B010105005, 2015A030311047, 201604046018).

\ifCLASSOPTIONcaptionsoff
  \newpage
\fi



\bibliographystyle{IEEEtran}
%



%
%


%

\begin{IEEEbiography}[{\includegraphics[width=0.8in,height=1in,clip,keepaspectratio]{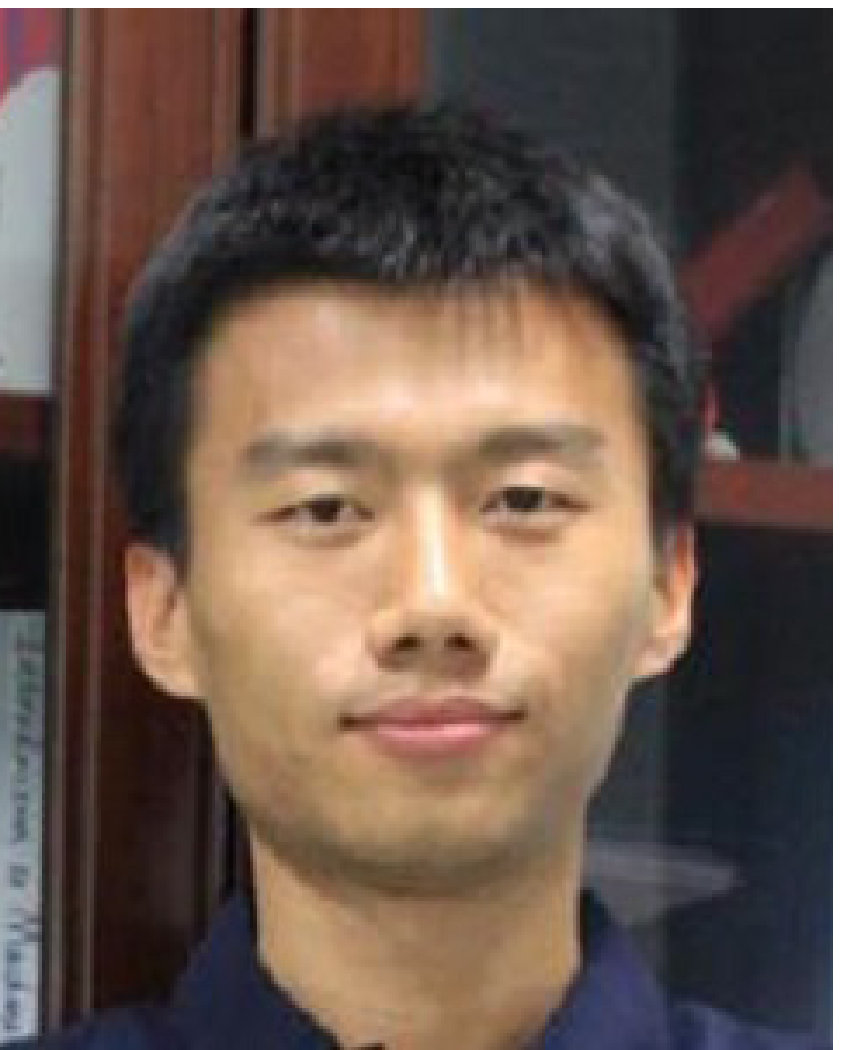}}]{Chunchao Guo}
received the B.E. degree in communication engineering with honors from Lanzhou University, China, in 2010. He is currently pursuing the Ph.D. degree in computer science at Sun Yat-sen University, China. His research interests are in computer vision and pattern recognition, with a focus on human identity recognition, object tracking, object detection and visual surveillance. Chun-Chao Guo is a recipient of the Excellent Paper Award at the 2014 National Conference on Image and Graphics. He won the first prize in the 2014 and 2015 National Graduate Contest on Smart-City Technology, and he was one of the winners in the 2014 Bocom Cup Contest on Video Analysis. He is a student member of CCF and IEEE.
\end{IEEEbiography}

\begin{IEEEbiography}[{\includegraphics[width=0.8in,height=1in,clip,keepaspectratio]{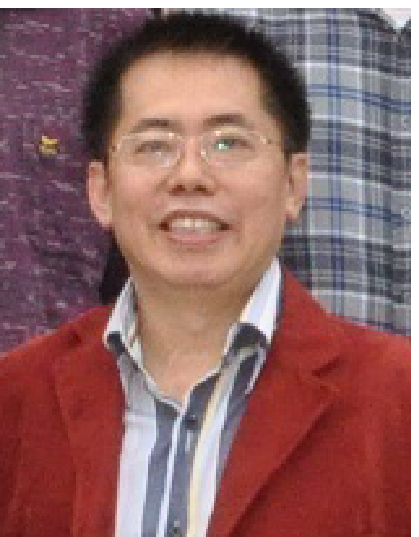}}]{Jianhuang Lai}
received his M.Sc. degree in applied mathematics in 1989 and his Ph.D. in mathematics in 1999 from SUN YAT-SEN University, China. He joined Sun Yat-sen University in 1989 as an Assistant Professor, where currently, he is a Professor in School of Data and Computer Science. His current research interests are in the areas of computer vision, pattern recognition and its applications. He has published over 250 scientific papers in the international journals and conferences on image processing and pattern recognition��e.g. IEEE TPAMI, IEEE TNN, IEEE TIP, IEEE TSMC (Part B), Pattern Recognition, ICCV, CVPR and ICDM. Prof. Lai serves as a deputy director of of the Image and Graphics Association of China and also serves as a standing director of the Image and Graphics Association of Guangdong. He is also the deputy director of Computer Vision Committee, China Computer Federation (CCF).
\end{IEEEbiography}


\begin{IEEEbiography}[{\includegraphics[width=0.8in,height=1in,clip,keepaspectratio]{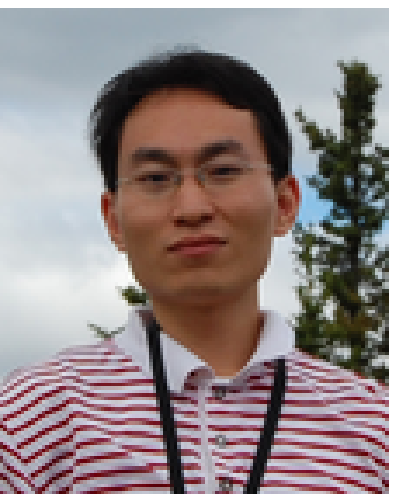}}]{Xiaohua Xie}
 is currently a Research Professor at Sun Yat-Sen University. Prior to joining SYSU, Xiaohua Xie was an Associate Professor at Shenzhen Institutes of Advanced Technology (SIAT), Chinese Academy of Sciences. He received the B.Sc. in Mathematics and Applied Mathematics (2005) from Shantou University, the M.Sc. in Information of Computing Science (2007) and the Ph.D. in Applied Mathematics (2010) from Sun Yat-sen University in China (jointly supervised by Concordia University in Canada). His current research fields cover image processing, computer vision, pattern recognition, and computer graphics, especially focusing on image understanding and object modeling. He has published more than a dozen papers in the prestigious international journals and conferences. He is recognized as Overseas High-Caliber Personnel (Level B) in Shenzhen, China.
\end{IEEEbiography}




\end{document}